\documentclass[10pt,twocolumn,letterpaper]{article}

\usepackage{cvpr}
\usepackage{times}
\usepackage{epsfig}
\usepackage{graphicx}
\usepackage{grffile}
\usepackage{amsmath}
\usepackage{amssymb}
\usepackage{multirow}
\usepackage{listings}
\usepackage{wasysym}
\usepackage[utf8]{inputenc}
\usepackage[dvipsnames]{xcolor}
\usepackage{color}
\usepackage{eso-pic}
\usepackage{graphicx}
\usepackage{array,booktabs,calc}

\usepackage[breaklinks=true,bookmarks=false]{hyperref}

\cvprfinalcopy % *** Uncomment this line for the final submission

\def\cvprPaperID{2340} % *** Enter the CVPR Paper ID here

\ifcvprfinal\fi
\begin{document}

%%%%%%%%% TITLE
\title{Neural Face Editing with Intrinsic Image Disentangling}

%Adversarial Face Editing\\Using Physically Based Render Disentangling for Neural Networks

\author{Zhixin Shu{\textsuperscript{1}} \hspace{1mm} Ersin Yumer{\textsuperscript{2}} \hspace{1mm} Sunil Hadap{\textsuperscript{2}} \hspace{1mm} Kalyan Sunkavalli{\textsuperscript{2}} \hspace{1mm} Eli Shechtman {\textsuperscript{2}} \hspace{1mm} Dimitris Samaras{\textsuperscript{1,3}}\\
	\\
\textsuperscript{1}Stony Brook University  \hspace{2mm} \textsuperscript{2}Adobe Research \hspace{2mm} \textsuperscript{3} CentraleSupélec, Université Paris-Saclay\\
\tt\small \textsuperscript{1}\{zhshu,samaras\}@cs.stonybrook.edu \hspace{2mm} \textsuperscript{2}\{yumer,hadap,sunkaval,elishe\}@adobe.com
}

\maketitle
%\thispagestyle{empty}

%%%%%%%%% ABSTRACT
\begin{abstract}
   Traditional face editing methods often require a number of sophisticated and task specific algorithms to be applied one after the other --- a process that is tedious, fragile, and computationally intensive. In this paper, we propose an end-to-end generative adversarial network that infers a face-specific disentangled representation of intrinsic face properties, including shape (i.e. normals), albedo, and lighting, and an alpha matte. We show that this network can be trained on ``in-the-wild'' images by incorporating an in-network physically-based image formation module and appropriate loss functions. Our disentangling latent representation allows for semantically relevant edits, where one aspect of facial appearance can be manipulated while keeping orthogonal properties fixed, and we demonstrate its use for a number of facial editing applications. 
\end{abstract}

%%%%%%%%% BODY TEXT
\section{Introduction}

Understanding and manipulating face images in-the-wild is of great interest to the vision and graphics community, and as a result, has been extensively studied in previous work. This ranges from techniques to relight portraits~\cite{wang2009face}, to edit or exaggerate expressions~\cite{yang2011expression}, and even drive facial performance~\cite{thies2016face}. Many of these methods start by explicitly reconstructing face attributes like geometry, texture, and illumination, and then edit these attributes to edit the image. However, reconstructing these attributes is a challenging and often ill-posed task; previous techniques deal with this by either assuming richer data (e.g., RGBD video streams) or a strong prior on the reconstruction that is adapted to the particular editing task that they seek to solve (e.g., low-dimensional geometry~\cite{blanz1999morphable}). As a result, these techniques tend to be both costly and not generalize well to the large variations in facial identity and appearance that exist in images-in-the-wild.

\begin{figure}
	\centering
	\begin{tabular}{c@{\hspace{0.02in}}c@{\hspace{0.02in}}c@{\hspace{0.02in}}c@{\hspace{0.02in}}c}		
		\includegraphics[height=0.6in]{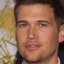} &
		\includegraphics[height=0.6in]{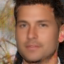} &
		\includegraphics[height=0.6in]{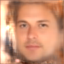} &
		\includegraphics[height=0.6in]{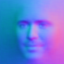} &
		\includegraphics[height=0.6in]{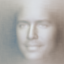} \\
		(a) input & (b) recon & (c) albedo & (d) normal & (e) shading \\
		\includegraphics[height=0.6in]{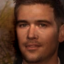}&
		\includegraphics[height=0.6in]{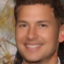} &
		\includegraphics[height=0.6in]{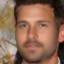} &
		\includegraphics[height=0.6in]{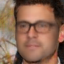} &
		\includegraphics[height=0.6in]{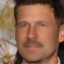} \\
		(f) relit & (g) smile & (h) beard & (i) eyewear & (j) older 
	\end{tabular}
	\caption{Given a face image (a), our network reconstructs the image (b) with in-network learned albedo (c), normal (d), and shading(e). Using this network, we can manipulate face through lighting (f), expression (g), appearance (h), eyewear (i), and time (j).}
	\label{fig:teaser}
	\vspace{-.1in}
\end{figure}

In this work, our goal is to learn a compact, meaningful manifold of facial appearance, and enable face edits by walking along paths on this manifold. The remarkable success of morphable face models~\cite{blanz1999morphable} -- where face geometry and texture are represented using low-dimensional linear manifolds -- indicates that this is possible for facial appearance. However, we would like to handle a much wider range of manipulations including changes in viewpoint, lighting, expression, and even higher-level attributes like facial hair and age -- aspects that cannot be represented using previous models. In addition, we would like to learn this model without the need for expensive data capture~\cite{cao2014facewarehouse}.

To this end, we build on the success of deep learning -- especially unsupervised auto-encoder networks -- to learn ``good'' representations from large amounts of data~\cite{bengio2013representation}. Trivially applying such approaches to our problem leads to representations that are not meaningful, making the subsequent editing challenging. However, we have (approximate) models for facial appearance in terms of \emph{intrinsic face properties} like geometry (surface normals), material properties (diffuse albedo), and illumination. We leverage this by designing the network to explicitly infer these properties and introducing an in-network forward rendering model that reconstructs the image from them. Merely introducing these factors into the network is not sufficient; because of the ill-posed nature of the inverse rendering problem, the learnt intrinsic properties can be arbitrary. We guide the network by imposing priors on each of these intrinsic properties; these include a morphable model-driven prior on the geometry, a Retinex-based~\cite{land1971lightness} prior on the albedo, and an assumption of low-frequency spherical harmonics-based lighting model~\cite{ramamoorthi2001relationship,basri2003lambertian}. By combining these constraints with adversarial supervision on image reconstruction, and weak supervision on the inferred face intrinsic properties, our network is able to learn disentangled representations of facial appearance.

Since we work with natural images, faces appear in front of arbitrary backgrounds, where the physical constraints of the face do not apply. Therefore, we also introduce a matte layer to separate the foreground (i.e., the face) from the image background. This enables us to provide optimal reconstruction pathways in the network specifically designed for faces, without distorting the background reconstruction.

Our network naturally exposes low-dimensional manifold embeddings for each of the intrinsic properties, which in turn enables direct and data-driven semantic editing from \emph{a single input image}. Specifically, we demonstrate direct illumination editing with explicit spherical harmonics lighting built into the network, as well as latent space manifold traversal for semantically meaningful expression edits such as smiling, and more structurally global edits such as aging. We show that by constraining physical properties that do not affect the target edits, we can achieve significantly more realistic results compared to other learning-based face editing approaches.

Our main contributions are:
(1) We introduce an end-to-end generative network specifically designed for the understanding and editing of face images in the wild;
(2) We encode the image formation and shading processes as in-network layers enabling the disentangling in the the latent space, of physically based rendering elements such as shape, illumination, and albedo;
(3) We introduce statistical loss functions (such as \emph{batchwise white shading} (BWS) corresponding to color consistency theory~\cite{land1971lightness}) to improve disentangling latent representations.

\section{Related Work}

\textbf{Face Image Manipulation.} Face modeling and editing is an extensively studied topic in vision and graphics. Blanz and Vetter~\cite{blanz1999morphable} showed that facial geometry and texture can be approximated by a low-dimensional morphable face model. This model and its variants have been used for a variety of tasks including relighting~\cite{wang2009face,chai2015portrait}, face attribute editing~\cite{cao2014facewarehouse}, expression editing~\cite{blanz2003reanimating,liu2001expressive}, authoring facial performances~\cite{vlasic2005multilinear,thies2016face}, and aging~\cite{kemelmacher2014illumination}. Another class of techniques uses coarse geometry estimates to drive image-based editing tasks~\cite{yang2011expression,shu2016eyes,kemelmacher2016transfiguring}. Each of these works develops techniques that are specifically designed for their application and often can not be generalized to other tasks. In contrast, our work aims to learn a general manifold for facial appearance that can support all these tasks.

\textbf{Intrinsic decompositions.} Barrow and Tanenbaum~\cite{Barrow78} proposed the concept of decomposing images into their physical intrinsic components such as surface normals, surface shading, etc. Barron and Mallik~\cite{barron2015shape} extended this decomposition
%to explicitly decompose an image into surface normals, albedo, and lighting. They 
assuming a Lambertian rendering model with low-frequency illumination and made use of extensive priors on geometry, albedo, and illumination. This rendering model has also been used in face relighting~\cite{wang2009face} and shape-from-shading-based face reconstruction~\cite{kemelmacher2011face}. We use a similar rendering model in our work, but learn a face-specific appearance model by training a deep network with weak supervision.

\textbf{Neural Inverse Rendering.} Generative network architectures have shown to be effective for image manipulation. Kulkarni et al.~\cite{dcign-2015} utilized a variational auto-encoder (VAE)~\cite{kingma2013auto} for synthesizing novel variations of the input image where the objects pose and lighting conditions are altered. Yang et al.~\cite{yang2015weakly} demonstrated novel view synthesis for the object in a given image, where view specific properties were disentangled in latent space utilizing a recurrent network. In contrast, Tatarchenko et al.~\cite{tatarchenko2016multi} used an autoencoder style network for the same task, where transformations were encoded through a secondary input stream and mixed with the input image in the latent space. Recently, Yan et al.~\cite{attribute2Image} used a VAE variant and layered representations to generate images with specific semantic attributes. We adopt their background-foreground disentangling scheme through an in-network matte layer. 

\textbf{Face Representation Learning.} Face representation learning is generally performed with a standard convolutional neural network trained for a recognition or labeling task~\cite{taigman2014deepface, parkhi2015deep, schroff2015facenet}. Such approaches often need a significantly large amount of data since the network is treated as a black box. Synthetically boosting the dataset using normalizations and augmentations~\cite{taigman2014deepface,hassner2013viewing,hassner2015effective} has proven useful. Most recently, Masi et al.~\cite{MTHLM:2016:dowe} used face fitting using morphable models similar to our approach, but used the resulting 3D faces to generate more data for traditional recognition network training. Even though such learned representations are powerful, especially in recognition, they are not straight forward to utilize for face editing. 

Recently, Gardner et al.~\cite{gardner2015deep} demonstrated face editing through a standard recognition network. Since the network does not have a natural generation pathway, they use a two step optimization procedure (one in latent space, and one at low level feature space) to reconstruct the edited image. This, combined with the fact that they use a  global latent space, leads to unintended changes and artifacts. On the other hand, our generative auto-encoder style network allows for a physically meaningful latent space disentangling, thereby solving both problems: we constrain semantic edits to their corresponding latent representation, and our decoder generates the editing result in a single forward pass.

\begin{figure*}
\centering
\includegraphics[width=0.94\textwidth]{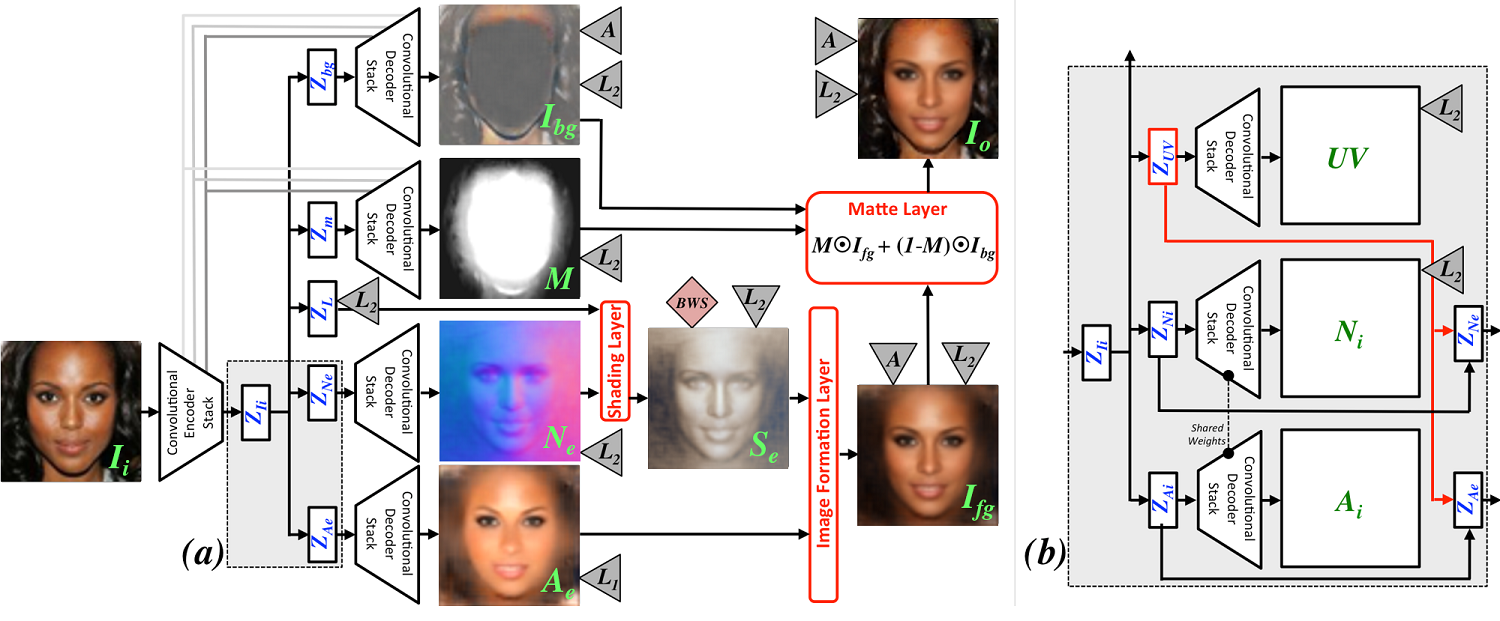}
\caption{\label{fig:network} Network Architectures. The interchangeable modules (grey background-dashed boundary) highlight the difference between our two proposed architectures: (a) Direct modeling of explicit normal ($N_e$) and albedo ($A_e$) maps. (b) Implicit coordinate system ($UV$), albedo ($A_i$) and normal ($N_i$) modeling to aid further disentangling in the face foreground.}
\end{figure*}

We formulate the face generation process as an end-to-end network where the face is physically grounded by explicit in-network representations of its shape, albedo, and lighting. Fig.~\ref{fig:network} shows the overall network structure. We first introduce the foreground \emph{Shading Layer} and the \emph{Image Formation Layer} (Sec.~\ref{sec:rendering}), followed by two alternative in-network face representations (Fig.~\ref{fig:network}(a)-(b) and Sec.~\ref{sec:representation}) that are compatible with in-network image formation. Finally, we  introduce in-network matting (Sec.~\ref{sec:matting}) which further disentangles the learning process of the foreground and background for face images in the wild.

\subsection{In-Network Physically-Based Face Rendering}\label{sec:rendering}

From a graphics point of view, we assume a given face image $I_{fg}$ is the result of a rendering process, $f_{\text{rendering}}$ where the inputs are an albedo map $A_e$, a normal map $N_e$, and illumination/lighting $L$:
\begin{equation}
	I_{fg}  = f_{\text{rendering}}(A_e,N_e,L)
	\label{equation:frendering}
\end{equation}
We assume Lambertian reflectance and adopt Retinex Theory~\cite{land1971lightness} to separate the albedo (i.e. reflectance) from the geometry and illumination:
\begin{equation}
I_{fg} = f_{\text{image-formation}}(A_e,S_e) = A_e \odot S_e
\label{equation:frenderI}
\end{equation}
in which $\odot$ denotes the per-element product operation in the image space, and $S_e$ represents a shading map rendered by $N_e$ and $L$:
\begin{equation}
S_e = f_{\text{shading}}(N_e,L)
\label{equation:frenderS}
\end{equation}

If Eqs.~\ref{equation:frenderI} and~\ref{equation:frenderS} are differentiable, they can be realized as in-network layers in an auto-encoder network (Fig.~\ref{fig:network}(a)). This allows us to represent the image with disentangled latent variables for physically meaningful factors in the image formation process: the albedo latent variable $Z_{A_e}$ , the normals variable $Z_{N_e}$ and the lighting variable $Z_L$. We show that this is advantageous over the traditional approach of a single latent variable that encodes the combined effect of all image formation factors. Each of the latent variables allows us access to a specific manifold, where semantically relevant edits can be performed while keeping irrelevant latent variables fixed. For instance, one can trivially perform image relighting by only traversing the lighting manifold given by $Z_L$ or changing only the albedo (e.g., to grow a beard) by traversing $Z_{A_e}$.

Computing shading from geometry ($N_e$) and illumination ($L$) is nontrivial under unconstrained conditions, and might result in $f_{\text{shading}}(\cdot,\cdot)$ being a discontinuous function in a significantly large region of the space it represents. Therefore, we further assume distant illumination, $L$, that is represented by {\em spherical harmonics}~\cite{ramamoorthi2001relationship} s.t. the Lambertian shading function, $f_{\text{shading}}(\cdot,\cdot)$ has an analytical form and is differentiable.

Following previous work~\cite{ramamoorthi2001relationship,basri2003lambertian,wang2009face,barron2015shape}, lighting $L$ is represented by a 9-dimensional spherical harmonics coefficient vector. For a given pixel, $i$, with normal $\mathbf{n_i} = [n_x, n_y, n_z]^{\top}$, the shading is rendered as:
\begin{equation}
S_e^i = S_e(\mathbf{n}_i, L) = [\mathbf{n_i};1]^{\top}K[\mathbf{n_i};1]
\label{equation:forward_shade}
\end{equation}
where
\begin{equation}
\begin{aligned}
K &= \left[ \begin{array}{cccc}
c_1 L_9 & c_1 L_5  & c_1 L_8 & c_2 L_4 \\
c_1 L_5 & -c_1 L_9 & c_1 L_6 & c_2 L_2 \\
c_1 L_8 & c_1 L_6  & c_3 L_7 & c_2 L_3 \\
c_2 L_4 & c_2 L_2  & c_2 L_3 & c_4 L_1 - c_5 L_7  \end{array} \right]
\\
c1 &= 0.429043 \quad c2 = 0.511664\\
c3 &= 0.743125
\quad c4 = 0.886227
\quad c5 = 0.247708\\
\end{aligned}
\end{equation}
We provide the formulas for the partial derivatives $\frac{\partial S_e^i}{\partial n_x}$, $\frac{\partial S_e^i}{\partial n_y}$,$\frac{\partial S_e^i}{\partial n_z}$ and $\frac{\partial S_e^i}{\partial L_j}$ in the supplementary material. Using these two differential rendering modules $f_{\text{shading}}$ and $f_{\text{image-formation}}$, we can now implement the rendering modules within the network as shown in Figure~\ref{fig:network}.

\subsection{In-Network Face Representation}\label{sec:representation}

\paragraph{Explicit Representation.} The formulation introduced in the previous section requires the image formation and shading variables to be defined in the image coordinate system. This can be achieved with an \emph{explicit} per-pixel representation of the face properties: $N_e$, $A_e$. Figure~\ref{fig:network}(a) depicts the module where the explicit normals and albedo are represented by their latent variables $Z_{N_e}$, $Z_{A_e}$. Note that the lighting, $L$, is independent of the face representation; we represent it using spherical harmonics coefficients, i.e., $Z_L=L$ is directly used by the shading layer whose forward process is given by Eqn.~\ref{equation:forward_shade}.

\paragraph{Implicit Representation.} Even though the explicit representation helps disentangle certain properties and relates edits more intuitively to the latent variable manifolds (i.e. relighting), it might not be satisfactory in some cases. For instance, pose and expression edits might change both the explicit per-pixel normals, as well as the per-pixel albedo in the image space. We therefore introduce an \emph{implicit} representation, where the parametrization is over the face coordinate system rather than the image coordinate system. This will allow us to further constrain pose and expression changes to the shape (i.e. normal) space only.

To address this, we introduce an alternative network architecture where the explicit representation depicted in the module in Fig.~\ref{fig:network}(a) is replaced with Fig.~\ref{fig:network}(b). Here, $UV$ represents the per-pixel face space uv-coordinates, $N_i$ and $A_i$ represent the normal and albedo maps in the face uv-coordinate system, and $Z_{UV}$, $Z_{N_i}$, and $Z_{A_i}$ represent the corresponding latent variables respectively. This is akin to the standard UV-mapping process in computer graphics. Facial features are aligned in this space (eyes correspond to eyes, mouths to mouths, etc.), and as a result the network has to learn a smaller space of variation, leading to sharper, more accurate reconstructions. Note that even though the network only uses the explicit latent variables at test time, we have auxiliary decoder stacks for all implicit variables to encourage disentangling of these variables during training. The implementation and training details will be explained in Sec.~\ref{section:training}.

\subsection{In-Network Background Matting}\label{sec:matting}

To further encourage the physically based representations of albedo, normals and lighting to concentrate on the face region, we disentangle the background from the foreground with a matte layer similar to the work by Yan et al.~\cite{attribute2Image}. The matte layer computes the compositing of the foreground face onto the background:
\begin{equation}
I_o = M \odot I_{fg}+ (1-M) \odot I_{bg}
\end{equation}
The matting layer also enables us to utilize efficient skip layers where unpooling layers in the decoder stack can use pooling switches from the corresponding encoder stack of the input image (grey links from the input encoder to background and mask decoders in Figure~\ref{fig:network}). The skip connection between the encoder and the decoder, allow for the details of the background to be preserved to a greater extent. Such skip connections bypass the bottleneck $Z$ and therefore allow only partial information flow through $Z$ during training.

For the foreground face region we chose to ``filter'' all the information through the bottleneck $Z$ without any skip connections in order to gain full control over the latent manifolds for editing, at the expense of some detail loss.

\section{Implementation}
\label{section:implementation}

\subsection{Network Architecture}
\label{section:architecture}

The convolutional encoder stack (Fig.~\ref{fig:network}) is composed of three convolutions with $32*3\times3$, $64*3\times3$ and $64*3\times3$ filter sets. Each convolution is followed by max-pooling and a ReLU nonlinearity. We pad the filter responses after each pooling layer so that the final output of the convolutional stack is a set of filter responses with size $64*8\times8$ for an input image $3*64\times64$. 

$Z_{I_i}$ is a latent variable vector of $128\times1$ which is fully connected to the last encoder stack downstream as well as the individual latent variables for background $Z_{bg}$, mask $Z_m$, light $Z_L$, and the foreground representations. For the explicit foreground representation, it is directly connected to $Z_{N_e}$ and $Z_{A_e}$ (Fig.~\ref{fig:network}(a)), whereas for the implicit representation it is connected to $Z_{UV}$, $Z_{N_i}$, and $Z_{A_i}$ (Fig.~\ref{fig:network}(b)). All individual latent representations are $128\times 1$ vectors except for $Z_L$ which represents the light $L$ directly and is thus a $27\times1$ vector (three $9\times1$ concatenated vectors representing the spherical harmonics of the RGB components). 

All decoder stacks for upsampling per-pixel (explicit or implicit) values are strictly symmetric to the encoder stack. As described in Sec.~\ref{sec:matting}, the decoder stacks for the mask and background have skip connections to the input encoder stack at corresponding layers. The implicit normals $N_i$ and implicit albedo $A_i$ share weights in the decoder, since we have supervision of the implicit normals only. 

\subsection{Training}
\label{section:training}

We use ``in-the-wild'' face images for training. Hence, we only have access to the image itself (denoted by $I^*$), and do not have ground-truth data for either illumination, normal map, or the albedo.
The main loss function is therefore on the reconstruction of the image $I_i$ at the output $I_o$:
\begin{equation}
E_o = E_\text{recon} + \lambda_\text{adv} E_\text{adv}
\end{equation}
where $E_\text{recon} = ||I_i- I_o||^2$. $E_\text{adv}$ is given by the adversarial loss, where a discriminative network is trained at the same time to distinguish between the generated and real images~\cite{goodfellow2014generative}. Specifically, we use an energy-based method~\cite{DBLP:journals/corr/ZhaoML16} to incorporate the adversarial loss. In this approach an autoencoder is used as the discriminative network, $\mathcal{D}$. The adversarial loss for the generative network is defined as $E_\text{adv} = D(I')$, where $I'$ is the reconstruction of the discriminator input $I_o$, hence $D(.)$ is the $L_2$ reconstruction loss of the discriminator $\mathcal{D}$. We train $\mathcal{D}$ to minimize the margin-based reconstruction error proposed by~\cite{DBLP:journals/corr/ZhaoML16}.

Fully unsupervised training using only the reconstruction and adversarial loss on the output image will often result in semantically meaningless latent representations. The network architecture itself cannot prevent degenerate solutions, e.g. when $A_e$ captures both albedo and shading information while $S_e$ remains constant. Since each of the rendering elements has a specific physical meaning, and they are explicitly encoded as intermediate layers in the network, we introduce additional constraints through intermediate loss functions to guide the training.

First, we introduce $\hat{N}$, a ``pseudo ground-truth'' of the normal map $N_e$, to keep the normal map close to plausible face normals during the training process. We estimate $\hat{N}$ by fitting coarse face geometry to every image in the training set using a 3D Morphable Model~\cite{blanz1999morphable}. We then introduce the following objective to $N_e$:
\begin{equation}
E_\text{recon-N} = ||N_e-\hat{N}||^2
\end{equation}

Similar to $\hat{N}$, we provide a $L2$ reconstruction loss w.r.t $\hat{L}$, on the lighting parameters $Z_L$:
\begin{equation}
E_\text{recon-L} = ||Z_L-\hat{L}||^2
\end{equation}
where $\hat{L}$ is computed from $\hat{N}$ and the input image using least square optimization and a constant albedo assumption~\cite{WangCVPR07,wang2009face}. 

Furthermore, following Retinex theory~\cite{land1971lightness} which assumes albedo to be piecewise constant and shading to be smooth, we introduce an $L1$ smoothness loss on the gradients of the albedo, $A$:
\begin{equation}
E_\text{smooth-A} =  ||\nabla A_e||
\end{equation}
in which $\nabla$ is the spatial image gradient operation. In addition, since the shading is assumed to vary smoothly, we introduce an $L2$ smoothness loss on the gradients of the shading, $S_e$:
\begin{equation}
E_\text{smoothS} =  ||\nabla S_e||^2
\end{equation}

For the implicit coordinate system ($UV$) variant (Fig.~\ref{fig:network}-(b)), we provide $L2$ supervisions to both $UV$ and $N_i$:
\begin{equation}
    E_\text{UV} = ||UV- \hat{UV}||^2
\end{equation}
\begin{equation}
    E_{N_i} = ||N_i- \hat{N_i}||^2
\end{equation}
$\hat{UV}$ and $\hat{N_i}$ are obtained from the previously mentioned Morphable Model, in which vertex-wise correspondence on the 3D fit exists. We utilize the average shape of the Morphable Model $\bar{S}$ to construct a canonical coordinate map ($UV$) and surface normal map ($N_i$), and propagate it to each shape estimation via this correspondence. More details of this computation are presented in our supplemental document.

Due to ambiguity in the magnitude of lighting, and therefore the intensity of shading (Eq. \ref{equation:frenderI}), it is necessary to incorporate constraints on the shading magnitude to prevent the network from generating arbitrary bright/dark shading. Moreover, since the illumination is separated in  individual colors $\mathbf{L}_r$, $\mathbf{L}_g$ and $\mathbf{L}_b$, we incorporate a constraint to prevent the shading from being too strong in one color channel vs. the others. To handle these ambiguities, we introduce a {\em Batch-wise White Shading (BWS)} constraint on $S_e$:
\begin{equation}
\frac{1}{m}\sum_{i,j}  s_r^i(j) = \frac{1}{m}\sum_{i,j} s_g^i(j) = \frac{1}{m}\sum_{i,j} s_b^i(j) = c
\end{equation}
where $s_r^i(j)$ denotes the $j$-th pixel of the $i$-th example in the first (red) channel of $S_e$. $s_g$ and $s_b$ denote the second and the third channel of shading respectively. $m$ is the number of pixels in a training batch. In all experiments $c = 0.75$.

Since $\hat{N}$ obtained by the Morphable Model comes with a region of interest only on the face surface, we use it as the mask under which we compute all foreground losses. In addition, this region of interest is also used as the mask pseudo ground truth at training time for learning the matte mask:
\begin{equation}
E_M = ||M - \hat{M}||^2
\end{equation}
in which $\hat{M}$ represents the Morphable Model mask.

\begin{figure*}
	\centering
	\begin{tabular}{c@{\hspace{0.01in}}c@{\hspace{0.02in}}c@{\hspace{0.02in}}c@{\hspace{0.02in}}c@{\hspace{0.02in}}c@{\hspace{0.02in}}c@{\hspace{0.02in}}c@{\hspace{0.02in}}c@{\hspace{0.02in}}c@{\hspace{0.02in}}c}
		
		\raisebox{2.5\height}{input} &
		\includegraphics[height=0.6in]{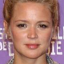} &
		\includegraphics[height=0.6in]{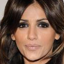} &
		\includegraphics[height=0.6in]{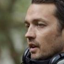} &
		\includegraphics[height=0.6in]{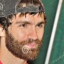} &
		\includegraphics[height=0.6in]{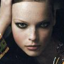} &
		\includegraphics[height=0.6in]{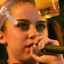} &
		\includegraphics[height=0.6in]{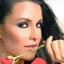} &
		\includegraphics[height=0.6in]{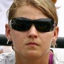} &
		\includegraphics[height=0.6in]{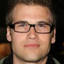} &
		\includegraphics[height=0.6in]{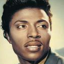} \\
		\raisebox{2.5\height}{baseline} &
		\includegraphics[height=0.6in]{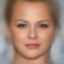} &
		\includegraphics[height=0.6in]{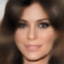} &
		\includegraphics[height=0.6in]{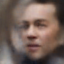} &
		\includegraphics[height=0.6in]{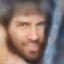} &
		\includegraphics[height=0.6in]{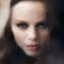} &
		\includegraphics[height=0.6in]{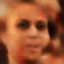} &
		\includegraphics[height=0.6in]{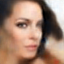} &
		\includegraphics[height=0.6in]{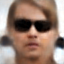} &
		\includegraphics[height=0.6in]{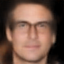} &
		\includegraphics[height=0.6in]{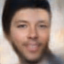} \\
		%\raisebox{2.5\height}{our recon} &
		\raisebox{3\height}{\vtop{\hbox{\strut our}\hbox{\strut recon}}}&
		\includegraphics[height=0.6in]{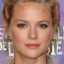} &
		\includegraphics[height=0.6in]{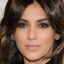} &
		\includegraphics[height=0.6in]{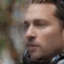} &
		\includegraphics[height=0.6in]{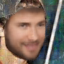} &
		\includegraphics[height=0.6in]{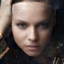} &
		\includegraphics[height=0.6in]{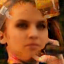} &
		\includegraphics[height=0.6in]{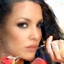} &
		\includegraphics[height=0.6in]{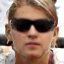} &
		\includegraphics[height=0.6in]{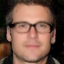} &
		\includegraphics[height=0.6in]{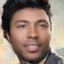} \\
		\raisebox{3\height}{\vtop{\hbox{\strut our}\hbox{\strut albedo}}}&
		\includegraphics[height=0.6in]{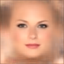} &
		\includegraphics[height=0.6in]{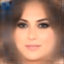} &
		\includegraphics[height=0.6in]{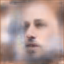} &
		\includegraphics[height=0.6in]{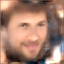} &
		\includegraphics[height=0.6in]{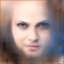} &
		\includegraphics[height=0.6in]{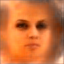} &
		\includegraphics[height=0.6in]{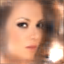} &
		\includegraphics[height=0.6in]{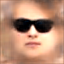} &
		\includegraphics[height=0.6in]{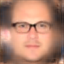} &
		\includegraphics[height=0.6in]{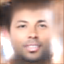} \\
		\raisebox{3\height}{\vtop{\hbox{\strut our}\hbox{\strut shading}}}&
		\includegraphics[height=0.6in]{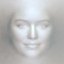} &
		\includegraphics[height=0.6in]{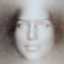} &
		\includegraphics[height=0.6in]{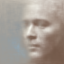} &
		\includegraphics[height=0.6in]{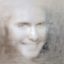} &
		\includegraphics[height=0.6in]{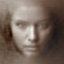} &
		\includegraphics[height=0.6in]{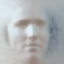} &
		\includegraphics[height=0.6in]{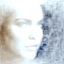} &
		\includegraphics[height=0.6in]{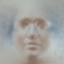} &
		\includegraphics[height=0.6in]{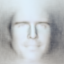} &
		\includegraphics[height=0.6in]{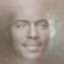} \\
		\raisebox{3\height}{\vtop{\hbox{\strut our}\hbox{\strut normal}}}&
		\includegraphics[height=0.6in]{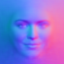} &
		\includegraphics[height=0.6in]{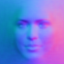} &
		\includegraphics[height=0.6in]{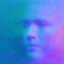} &
		\includegraphics[height=0.6in]{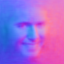} &
		\includegraphics[height=0.6in]{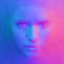} &
		\includegraphics[height=0.6in]{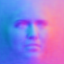} &
		\includegraphics[height=0.6in]{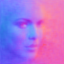} &
		\includegraphics[height=0.6in]{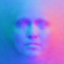} &
		\includegraphics[height=0.6in]{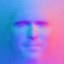} &
		\includegraphics[height=0.6in]{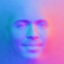} \\
		\raisebox{2.5\height}{3dMM} &
		\includegraphics[height=0.6in]{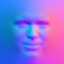} &
		\includegraphics[height=0.6in]{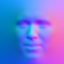} &
		\includegraphics[height=0.6in]{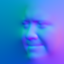} &
		\includegraphics[height=0.6in]{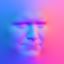} &
		\includegraphics[height=0.6in]{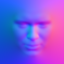} &
		\includegraphics[height=0.6in]{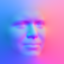} &
		\includegraphics[height=0.6in]{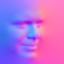} &
		\includegraphics[height=0.6in]{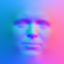} &
		\includegraphics[height=0.6in]{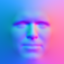} &
		\includegraphics[height=0.6in]{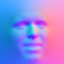} \\

		 & (1) & (2) & (3) & (4) & (5) & (6) & (7) & (8) & (9) & (10)
 	\end{tabular}
	\caption{Feedforward reconstruction and normals, shading, albedo estimation. Compared to the baseline auto-encoder (row 2), our reconstruction (row 3) not only preserves the details of the background (1,2,4), but is also more robust to complex pose (3,4), illumination (5), and identity (9,10), thanks to the layered representation and in-network rendering procedure. Moreover, our network contains components that explicitly encode normals (row 4), shading/lighting (row 5), and albedo (row 6) for the foreground (face), which is helpful for the understanding and manipulation of face images. In the last row we show the normal estimation from a 3D Morphable Model. We can easily see that using our network, the generated shape retains more identity information from the original image, and does not fall in the sub-space of the PCA-based morphable model that is used as weak supervision for training. All results are produced by the network with architecture designed for explicit face representation (Figure~\protect\ref{fig:network}(a)).}
	\label{fig:cmp_reconstruction}
\end{figure*}

\section{Experiments}

We use the CelebA~\cite{liu2015deep} dataset to train the network. For each image in the dataset, we detect landmarks~\cite{saragih2011principal}, and fit a 3D Morphable Model~\cite{blanz1999morphable,yang2011expression} to the face region to have a rough estimation of the rendering elements ($\hat{N}, \hat{L}$). These estimates are used to set-up the various losses detailed in the previous section. This data is subsequently used only for the training of the network as previously described.

\subsection{Baseline Comparisons}

For comparison, we train an auto-encoder $\mathcal{B}$ as a baseline. The encoder and decoder of $\mathcal{B}$ is identical to the encoder and decoder for albedo in our architecture. To make the comparison fair, the bottleneck layer of $\mathcal{B}$ is set to $265$ ($=128+128+9$) dimensions, which is more than twice as large as the bottleneck layer in our architecture (size $128$), yielding slightly more capacity for the baseline. 
Even though our architecture has a narrower bottleneck, the disentangling of the latent factors and the presence of physically based rendering layers, lead to reconstructions that are more robust to complex background, pose, illumination, occlusion, etc.,~(Fig. \ref{fig:cmp_reconstruction}).

More importantly, given an input face images, our network provides explicit access to an estimation of the albedo, shading and normal map (Fig.~\ref{fig:cmp_reconstruction}) for the face. Notably, in the last row of Fig.~\ref{fig:cmp_reconstruction}, we compare the inferred normals from our network with the normals estimated from the input image using the 3D morphable model that we deployed to guide the training process. The data to construct the morphable model contains only 16 identities; this small subspace of identity variation leads to normals that are often inaccurate approximations of the true face shape (row 7 in Fig.~\ref{fig:cmp_reconstruction}). By using these estimates as weak supervision in combination with an appearance-based rendering loss, our network is able to generate normal maps (row 6 in Fig.~\ref{fig:cmp_reconstruction}) that extend beyond the morphable model subspace, better fit the shape of the input face, and exhibit more identity information. Please refer to our supplementary material for more comparisons.

\subsection{Face Editing by Manifold Traversal}

\begin{figure}
	\centering
	\begin{tabular}{c@{\hspace{0.05in}}c@{\hspace{0.05in}}c@{\hspace{0.05in}}c@{\hspace{0.05in}}c}
		
		\includegraphics[height=0.6in]{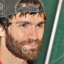} &
		\includegraphics[height=0.6in]{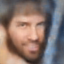} &
		\includegraphics[height=0.6in]{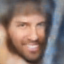} &
		\includegraphics[height=0.6in]{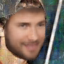} &
		\includegraphics[height=0.6in]{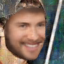} \\
		
		\includegraphics[height=0.6in]{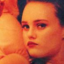} &
		\includegraphics[height=0.6in]{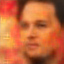} &
		\includegraphics[height=0.6in]{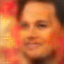} &
		\includegraphics[height=0.6in]{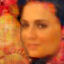} &
		\includegraphics[height=0.6in]{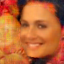} \\
		
		(a) & (b)  & (c) & (d) & (e)  
	\end{tabular}
	\caption{Smile editing via traversal on our representation (explicit albedo and normal) vs. a baseline auto-encoder representation. Our network provides better reconstructions (d)  of the input images (a) and captures the geometry and appearance changes associated with smiling (e). The baseline network leads to poorer reconstructions (b) and edits (c).}
	\label{fig:smile}
\end{figure}

Our network enables manipulation of semantic face attributes, (e.g. expression, facial hair, age, makeup, and eye-wear) by traversing the manifold(s) of the disentangled latent spaces that are most appropriate for that edit.

\begin{figure}[b]
	\centering
	\begin{tabular}{c@{\hspace{0.2in}}c@{\hspace{0.2in}}c}
		
		\includegraphics[height=0.8in]{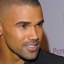} &
		\includegraphics[height=0.8in]{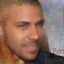} & 
		\includegraphics[height=0.8in]{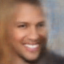}\\
		(a) input & (b) reconstruction & (c) baseline\\
		\includegraphics[height=0.8in]{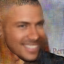} &
		\includegraphics[height=0.8in]{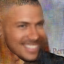} &
		\includegraphics[height=0.8in]{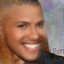} \\
		 (d)$Z_{UV}$ & (e) $Z_{UV}$,$Z_{Ni}$  & (f)  $Z_{UV}$,$Z_{Ni}$,$Z_{Ai}$
	\end{tabular}
	\caption{Smile editing via implicit factor traversal. Our implicit representation directly captures smiling via a traversal of the UV manifold (d) and both UV and implicit normals (e). Traversing on the implict albedos on the other hand, leads to noticeable appearance artifacts (f). For this experiment, we use the same regularization ($\lambda$= 0.03) on all manifolds.}
	\label{fig:uv_smile}
\end{figure}

For a given attribute, e.g., the \textit{smiling expression}, we feed both positive data $\{\mathbf{x}_p\}$ (smiling faces) and negative data $\{\mathbf{x}_n\}$ (faces with other expressions) into our network to generate two sets of $Z$-codes $\{\mathbf{z}_p\}$ and $\{\mathbf{z}_n\}$. These sets represent corresponding empirical distributions of the data on the low dimensional $Z$-space(s). Given an input face image $I_\text{source}$ that is \textit{not smiling}, we seek to \textit{make it smile} by moving its $Z$-code(s) $Z_\text{source}$ towards the distribution $\{\mathbf{z}\}_p$ to get a transformed code $Z_\text{trans}$. After that, we reconstruct the image corresponding to $Z_\text{trans}$ with the decoders in our model.

In order to compute the distributions for each attribute, we sample a subset of 2000 images from the CelebA~\cite{liu2015deep} with the appropriate attribute label (e.g., smiling vs other expresssions). We use the manifold traversal method proposed by Gardner et al.~\cite{gardner2015deep} independently on each appropriate variable. The extent of the traversal is parameterized by a regularization parameter, $\lambda$ (see~\cite{gardner2015deep} for details).  

In Fig.~\ref{fig:smile}, we compare the results using our network against the baseline auto-encoder. We traverse the albedo and normal variables to produce edits which make the faces smile and are able to capture changes in expression and the appearance of teeth, while preserving the other aspects of the image. In contrast, the results from traversing the baseline latent space are much poorer -- in addition to not being able to reconstruct the pose and identity of the input properly, the traversal is not able to capture the smiling transformation as well as we do.

\begin{figure}
	\centering
	\begin{tabular}{c@{\hspace{0.05in}}c@{\hspace{0.05in}}c@{\hspace{0.05in}}c@{\hspace{0.05in}}c}
		
		\includegraphics[height=0.6in]{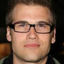} &
		\includegraphics[height=0.6in]{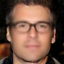} &
		\includegraphics[height=0.6in]{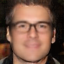} &
		\includegraphics[height=0.6in]{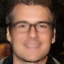} &
		\includegraphics[height=0.6in]{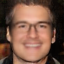} \\
		
		\includegraphics[height=0.6in]{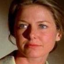} &
		\includegraphics[height=0.6in]{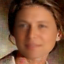} &
		\includegraphics[height=0.6in]{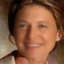} &
		\includegraphics[height=0.6in]{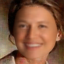} &
		\includegraphics[height=0.6in]{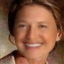} \\
		
		\includegraphics[height=0.6in]{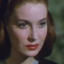} &
		\includegraphics[height=0.6in]{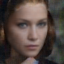} &
		\includegraphics[height=0.6in]{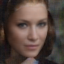} &
		\includegraphics[height=0.6in]{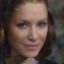} &
		\includegraphics[height=0.6in]{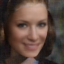} \\
		
		\includegraphics[height=0.6in]{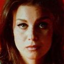} &
		\includegraphics[height=0.6in]{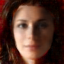} &
		\includegraphics[height=0.6in]{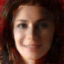} &
		\includegraphics[height=0.6in]{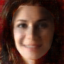} &
		\includegraphics[height=0.6in]{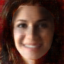} \\
		
	    \includegraphics[height=0.6in]{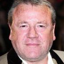} &
		\includegraphics[height=0.6in]{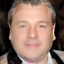} &
		\includegraphics[height=0.6in]{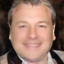} &
		\includegraphics[height=0.6in]{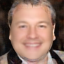} &
		\includegraphics[height=0.6in]{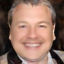} \\
		
		(a) input & (b) recon & (c)  & (d)  & (e) 
	\end{tabular}
	\caption{\textit{Smile} editing via progressive traversal on the bottleneck manifolds ($Z_{UV}$ and $Z_{N_i}$). From (c) to (e), $\lambda$ is $0.07$, $0.05$, $0.03$ respectively.  As the latent representation moves closer to the $smiling$ mode, stronger features of smiling, such as rising cheeks and white teeth, appear. Note that we are also able to capture subtle changes in the eyes that are often correlated with smiling.}
	\label{fig:smile2}
\end{figure}

In Fig.~\ref{fig:uv_smile} we demonstrate the utility of our implicit representation. While lips/mouth and teeth might map to the same region of the image space, they are in fact separated in the face UV-space. This allows the implicit variables to learn more targeted and accurate representations, hence traversing just the $Z_{UV}$, already results in a smiling face. Combining this with traversal along $Z_{N_i}$ exaggerates the smile. In contrast, we do not expect smiling to be correlated with the implicit albedo space, and traversing along the $Z_{A_i}$ leads to poorer results with an incorrect frontal pose.

In Fig.~\ref{fig:smile2} we demonstrate more results for \textit{smiling} and demonstrate that relaxing the traversal regularization parameter, $\lambda$, gradually leads to stronger smiling expressions.

\begin{figure}
	\centering
	\begin{tabular}{c@{\hspace{0.05in}}c@{\hspace{0.05in}}c@{\hspace{0.05in}}c@{\hspace{0.05in}}c}
		
		\includegraphics[height=0.6in]{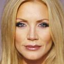} &
		\includegraphics[height=0.6in]{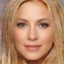} &
		\includegraphics[height=0.6in]{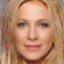} &
		\includegraphics[height=0.6in]{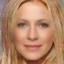} &
		\includegraphics[height=0.6in]{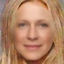} \\

		\includegraphics[height=0.6in]{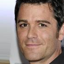} &
		\includegraphics[height=0.6in]{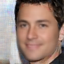} &
		\includegraphics[height=0.6in]{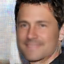} &
		\includegraphics[height=0.6in]{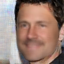} &
		\includegraphics[height=0.6in]{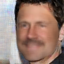} \\
			
	    \includegraphics[height=0.6in]{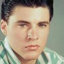} &
		\includegraphics[height=0.6in]{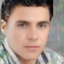} &
		\includegraphics[height=0.6in]{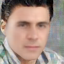} &
		\includegraphics[height=0.6in]{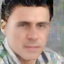} &
		\includegraphics[height=0.6in]{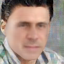} \\
		
		\includegraphics[height=0.6in]{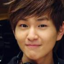} &
		\includegraphics[height=0.6in]{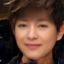} &
		\includegraphics[height=0.6in]{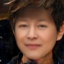} &
		\includegraphics[height=0.6in]{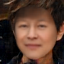} &
		\includegraphics[height=0.6in]{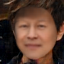} \\
		
		\includegraphics[height=0.6in]{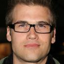} &
		\includegraphics[height=0.6in]{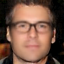} &
		\includegraphics[height=0.6in]{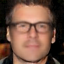} &
		\includegraphics[height=0.6in]{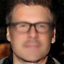} &
		\includegraphics[height=0.6in]{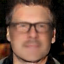} \\
		(a) input & (b) recon  & (c) & (d) & (e)
	\end{tabular}
	\caption{Aging via traversal on the albedo and normal manifolds. From (c) to (e), $\lambda$ is $0.07$, $0.05$, $0.03$ respectively. As the latent representation moving towards to the \textit{senior} mode, stronger features of \textit{aging}, such as changes in face shape and texture, appear while retaining other aspects of the appearance like pose, lighting, and eyewear.}
	\label{fig:age}
\end{figure}

We also address the editing task of \textit{aging} via manifold traversal. For this experiment, we construct the latent space distributions using images and labels from the PubFig~\cite{kumar2009attribute} dataset corresponding to the most and least \textit{senior} images. We expect aging to be correlated with both shape and texture, and show in Fig.~\ref{fig:age} that traversing these manifolds leads to convincing age progression.

Note that all of these edits have been performed on the exact same network, indicating that our network architecture is general enough to represent the manifold of face appearance, and is able to disentangle the latent factors to support specific editing tasks. Refer to our supplementary material for more results, and comparisons.

\textbf{Limitations.} Our current face masks do not include hair. This results in less control over some edits, e.g. aging, that are inherently affecting the hair as well. However, this can trivially be addressed, if a mask that also includes the hair can be generated ~\cite{chai2016autohair}.

\subsection{Relighting}
A direct application of the albedo-normal-light decomposition in our network is that it allows us to manipulate the illumination of an input face via $Z_L$ while keeping the other latent variable fixed. We can directly ``relight'' the face by replacing its $Z_L^{\text{target}}$ with some other $Z_L^{\text{source}}$ (e.g. using the lighting variable of another face).  

\begin{figure}
	\centering
	\begin{tabular}{c@{\hspace{0.05in}}c@{\hspace{0.05in}}c@{\hspace{0.05in}}c@{\hspace{0.05in}}c}
		
		\includegraphics[height=0.6in]{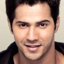} &
		\includegraphics[height=0.6in]{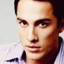} &
		\includegraphics[height=0.6in]{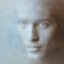} &
		\includegraphics[height=0.6in]{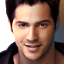} &
		\includegraphics[height=0.6in]{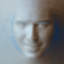} \\
		\includegraphics[height=0.6in]{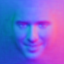}&
	    \includegraphics[height=0.6in]{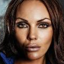} &
		\includegraphics[height=0.6in]{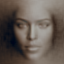} &
		\includegraphics[height=0.6in]{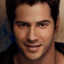} &
		\includegraphics[height=0.6in]{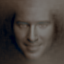} \\
		(a) target & (b) source & (c) $S^{\text{source}}$ & (d) transfer & (e) $S^{\text{transfer}}$ 
	\end{tabular}
	\caption{Lighting transfer using our model. We transfer the illumination of two source images (b) to a given target (a)(top: image; bottom: estimated normal), by generating the shading (e) of the target using the lighting of the source, and applying to the original target image.}
	\label{fig:lighting_transfer}
\end{figure}

While our network is trained to reconstruct the input, due to its limited capacity (especially due to the bottleneck layer dimensionality), the reconstruction does not reproduce the input with all the details. For illumination editing, however, we can directly manipulate the shading, that is also available in our network. We pass the source $I^{\text{source}}$ and target images $I^{\text{target}}$ through our network to estimate their individual factors. We use the target shading $S^{\text{target}}$ with  Eq.~\ref{equation:frenderI} to compute a ``detailed'' albedo $A^{\text{target}}$. Given the source light $L^{\text{source}}$, we render the shading of the target under this light with the target normals $N^{\text{target}}$ (Eq.~\ref{equation:frenderS}) to obtain the transferred shading $S^{\text{transfer}}$. In the end, the lighting transferred image is rendered with $A^{\text{target}}$ and $S^{\text{transfer}}$ using Eq.~\ref{equation:frenderI}. This is demonstrated in Fig.~\ref{fig:lighting_transfer} where we are able to successfully transfer the lighting from two sources with disparate identities, genders, and poses to a target while retaining all its details. We present more relighting results, as well as quantitative tests on illumination (i.e. spherical harmonics coefficients) prediction in the supplementary material. 

\section{Conclusions}

We proposed a physically grounded rendering-based disentangling network specifically designed for faces. Such disentangling enables realistic face editing since it allows trivial constraints at manipulation time. We are the first to attempt in-network rendering for faces in the wild with real, arbitrary backgrounds. Comparisons with traditional auto-encoder approaches show significant improvements on final edits, and our intermediate outputs such as face normals show superior identity preservation compared to traditional approaches.

\section{Acknowledgements}
This work started when Zhixin Shu was an intern at Adobe Research. This work was supported by a gift from Adobe, NSF IIS-1161876, the Stony Brook SensorCAT and the Partner University Fund 4DVision project.

{\small
\bibliographystyle{ieee}
\bibliography{camera_ready_small}
}
\clearpage
\appendix

\section{Implementation: more details}

In this section, we provide more details regarding the implementation of the rendering layers $f_\text{shading}$ and $f_\text{image-formation}$ as described in the paper.

\subsection{Shading Layer}

The shading layer is rendered with a spherical harmonics illumination representation~\cite{ramamoorthi2001relationship,basri2003lambertian,wang2009face,barron2015shape}.

The forward process is described by equations (3),(4), and (5) in the main paper.
We now provide the backward process, i.e., the partial derivatives $\frac{\partial S_e^i}{\partial n_x}$, $\frac{\partial S_e^i}{\partial n_y}$,$\frac{\partial S_e^i}{\partial n_z}$ and $\frac{\partial S_e^i}{\partial L_j}$ as follows:
\begin{equation}
	\frac{\partial S_e^i}{\partial n_x} = 2(c_1L_9n_x+c_1L_5n_y+c_1L_8n_z+c_2L_4)
\end{equation}
\begin{equation}
	\frac{\partial S_e^i}{\partial n_y} = 2(c_1L_5n_x-c_1L_9n_y+c_1L_6n_z+c_2L_2)
\end{equation}
\begin{equation}
	\frac{\partial S_e^i}{\partial n_z} = 2(c_1L_8n_x+c_1L_6n_y+c_3L_7n_z+c_2L_3)
\end{equation}
\begin{equation}
\frac{\partial S_e^i}{\partial L_1} = c_4
\end{equation}
\begin{equation}
\frac{\partial S_e^i}{\partial L_2} = 2c_2n_y
\end{equation}
\begin{equation}
\frac{\partial S_e^i}{\partial L_3} = 2c_2n_z
\end{equation}
\begin{equation}
\frac{\partial S_e^i}{\partial L_4} = 2c_2n_x
\end{equation}
\begin{equation}
\frac{\partial S_e^i}{\partial L_5} = 2c_1n_xn_y
\end{equation}
\begin{equation}
\frac{\partial S_e^i}{\partial L_6} = 2c_1n_yn_z
\end{equation}
\begin{equation}
\frac{\partial S_e^i}{\partial L_7} = c_3n_z^2-c_5
\end{equation}
\begin{equation}
\frac{\partial S_e^i}{\partial L_8} = 2c_1n_xn_z
\end{equation}
\begin{equation}
\frac{\partial S_e^i}{\partial L_9} = c_1n_x^2 - c_1n_y^2
\end{equation}
where
\begin{equation}
\begin{aligned}
c1 &= 0.429043 \quad c2 = 0.511664\\
c3 &= 0.743125
\quad c4 = 0.886227
\quad c5 = 0.247708\\
\end{aligned}
\end{equation}
\subsection{Image Formation Layer}

The forward process of the image formation layer (for the foreground) is simply a per-element product (see equation (2) in the main paper), therefore the backward process (partial derivatives) is:
\begin{equation}
	\frac{\partial I_{fg}}{\partial A_e^i} = S_e^i 
\end{equation} 
and 
\begin{equation}
\frac{\partial I_{fg}}{\partial S_e^i} = A_e^i 
\end{equation} 

\section{Quantitative Experiments}

\begin{figure}
	\centering
	\includegraphics[height=2.6in]{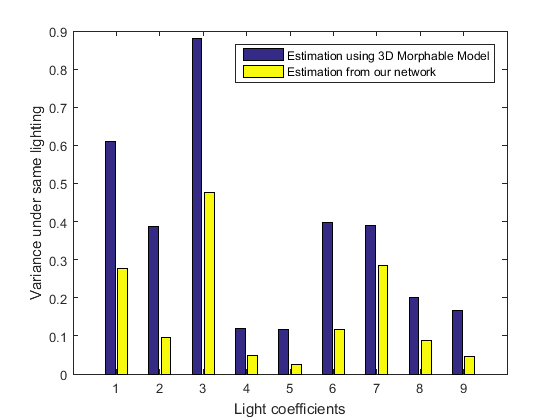}
	\caption{Stability of illumination estimation using our network. The illumination estimation from our network in the form of spherical harmonics lighting coefficients is more stable, i.e., exhibits less variance while measuring illumination of face images under identical lighting in Multi-PIE~\cite{gross2010multi}, than the estimation from a least squares morphable model. }
	\label{fig:light_variance}
\end{figure}

%\begin{comment}
In order to evaluate the illumination estimation of our network, we utilize the Multi-PIE~\cite{gross2010multi} dataset where controlled illumination is available\footnote{Controlled illumination in this context means that the different subjects have been illuminated under the same lighting conditions. Hence, the correspondences across subjects for the same lighting is known, but the actual lighting setup, or conditions are not known.}. We randomly sample 7,000 images with different identities and poses under 20 controlled light sources. We measure the variance of illumination coefficients $L_i$ ($i= 1,...,9$) within an illumination condition. The average variance of a 3D Morphable Model with least square estimation is $0.36$ while the average variance of our network is $0.16$. In Fig.~\ref{fig:light_variance}, we show the average (among 20 lighting conditions and RGB) variance of each lighting coefficient. Note that our model is only trained with CelebA dataset while the images from the Multi-PIE datasets are only used for quantitative evaluation.

\begin{figure}
	\centering
	\begin{tabular}{c@{\hspace{0.01in}}c@{\hspace{0.02in}}c@{\hspace{0.02in}}c@{\hspace{0.02in}}c@{\hspace{0.02in}}c}
		(a) &
		\includegraphics[height=0.5in]{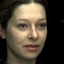} &
		\includegraphics[height=0.5in]{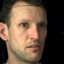} &
		\includegraphics[height=0.5in]{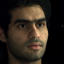} &
		\includegraphics[height=0.5in]{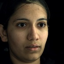} &
		\includegraphics[height=0.5in]{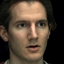} \\
		(b)&
		\includegraphics[height=0.5in]{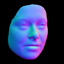} &
		\includegraphics[height=0.5in]{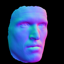} &
		\includegraphics[height=0.5in]{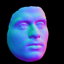} &
		\includegraphics[height=0.5in]{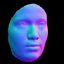} &
		\includegraphics[height=0.5in]{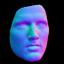} \\ ~\\
		(c)&
		\includegraphics[height=0.5in]{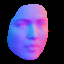} &
		\includegraphics[height=0.5in]{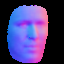} &
		\includegraphics[height=0.5in]{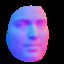} &
		\includegraphics[height=0.5in]{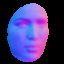} &
		\includegraphics[height=0.5in]{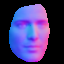} \\
		(d)&
		\includegraphics[height=0.5in]{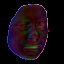} &
		\includegraphics[height=0.5in]{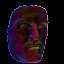} &
		\includegraphics[height=0.5in]{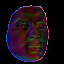} &
		\includegraphics[height=0.5in]{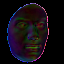} &
		\includegraphics[height=0.5in]{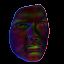} \\
		 error & 0.14 & 0.15  & 0.16 & 0.13  & 0.14 \\~\\
		(e)&
		\includegraphics[height=0.5in]{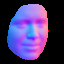} &
		\includegraphics[height=0.5in]{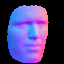} &
		\includegraphics[height=0.5in]{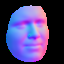} &
		\includegraphics[height=0.5in]{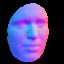} &
		\includegraphics[height=0.5in]{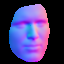} \\
		(f)&
		\includegraphics[height=0.5in]{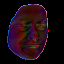} &
		\includegraphics[height=0.5in]{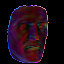} &
		\includegraphics[height=0.5in]{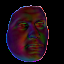} &
		\includegraphics[height=0.5in]{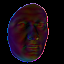} &
		\includegraphics[height=0.5in]{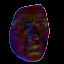} \\
		error & 0.16 & 0.16  & 0.17 & 0.15  & 0.15 
	\end{tabular}
	\caption{Comparison of normal reconstruction to ground truth. Input images (a), and ground truth normals (b). Our reconstruction and error (c,d). 3DMM fit and error (e,f) }
	\label{fig:supp_gtnormal}
\end{figure}

%\end{comment}
We evaluate the quality of our normal reconstruction vs. a direct 3D Morphable Model (3DMM) fit. We create input images for five different individuals (two women, three men) using light stage data captured by Weyrich et al.~\cite{weyrich2006faces}. We fit a 3DMM to these images and compute normals. We also pass these image as inputs to our disentangling network to get a normal reconstruction. In Fig.~\ref{fig:supp_gtnormal}, we compare the ground truth normals to our estimates and the 3DMM normals. Even though our network was trained on normals from morphable model fits, the additional reconstruction losses enable it to expand beyond this subspace. This is especially apparent for the reconstruction of the two women, whose faces look man-like in the 3DMM fits. %The error of our reconstruction is also lower at XXX degrees vs YYY degrees for 3DMM. 

\section{Additional Results}

We present more  face editing results, %using the manifold traversal method proposed by Gardner et al.~\cite{gardner2015deep} (as described in section 5.2 in the main paper), 
as well as relighting results using our proposed method described in section 5.3 of the main paper.

\begin{figure*}
	\centering
	\begin{tabular}{c@{\hspace{0.01in}}c@{\hspace{0.02in}}c@{\hspace{0.02in}}c@{\hspace{0.02in}}c@{\hspace{0.02in}}c@{\hspace{0.02in}}c@{\hspace{0.02in}}c@{\hspace{0.02in}}c@{\hspace{0.02in}}c}
		input &
\includegraphics[height=0.6in]{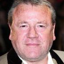} &
\includegraphics[height=0.6in]{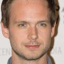} &
\includegraphics[height=0.6in]{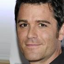} &
\includegraphics[height=0.6in]{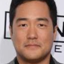} &
\includegraphics[height=0.6in]{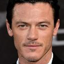} &
\includegraphics[height=0.6in]{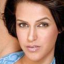} &
\includegraphics[height=0.6in]{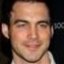} &
\includegraphics[height=0.6in]{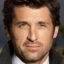} &
\includegraphics[height=0.6in]{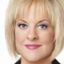} \\
recon &
\includegraphics[height=0.6in]{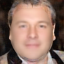} &
\includegraphics[height=0.6in]{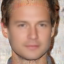} &
\includegraphics[height=0.6in]{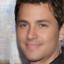} &
\includegraphics[height=0.6in]{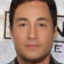} &
\includegraphics[height=0.6in]{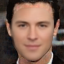} &
\includegraphics[height=0.6in]{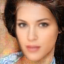} &
\includegraphics[height=0.6in]{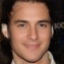} &
\includegraphics[height=0.6in]{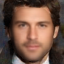} &
\includegraphics[height=0.6in]{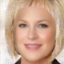} \\
result &
\includegraphics[height=0.6in]{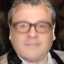} &
\includegraphics[height=0.6in]{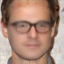} &
\includegraphics[height=0.6in]{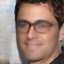} &
\includegraphics[height=0.6in]{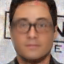} &
\includegraphics[height=0.6in]{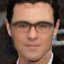} &
\includegraphics[height=0.6in]{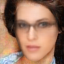} &
\includegraphics[height=0.6in]{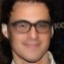} &
\includegraphics[height=0.6in]{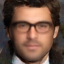} &
\includegraphics[height=0.6in]{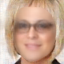} \\~\\

input &
\includegraphics[height=0.6in]{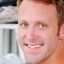} &
\includegraphics[height=0.6in]{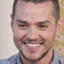} &
\includegraphics[height=0.6in]{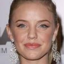} &
\includegraphics[height=0.6in]{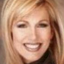} &
\includegraphics[height=0.6in]{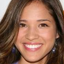} &
\includegraphics[height=0.6in]{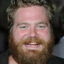} &
\includegraphics[height=0.6in]{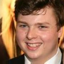} &
\includegraphics[height=0.6in]{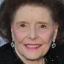} &
\includegraphics[height=0.6in]{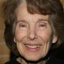} \\
recon &
\includegraphics[height=0.6in]{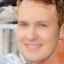} &
\includegraphics[height=0.6in]{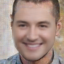} &
\includegraphics[height=0.6in]{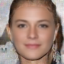} &
\includegraphics[height=0.6in]{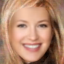} &
\includegraphics[height=0.6in]{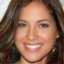} &
\includegraphics[height=0.6in]{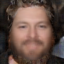} &
\includegraphics[height=0.6in]{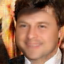} &
\includegraphics[height=0.6in]{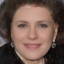} &
\includegraphics[height=0.6in]{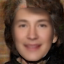} \\
result &
\includegraphics[height=0.6in]{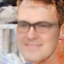} &
\includegraphics[height=0.6in]{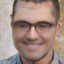} &
\includegraphics[height=0.6in]{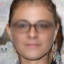} &
\includegraphics[height=0.6in]{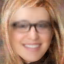} &
\includegraphics[height=0.6in]{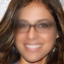} &
\includegraphics[height=0.6in]{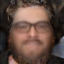} &
\includegraphics[height=0.6in]{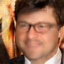} &
\includegraphics[height=0.6in]{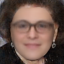} &
\includegraphics[height=0.6in]{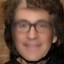} \\~\\

input &
\includegraphics[height=0.6in]{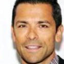} &
\includegraphics[height=0.6in]{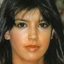} &
\includegraphics[height=0.6in]{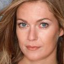} &
\includegraphics[height=0.6in]{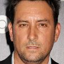} &
\includegraphics[height=0.6in]{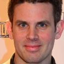} &
\includegraphics[height=0.6in]{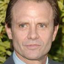} &
\includegraphics[height=0.6in]{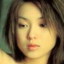} &
\includegraphics[height=0.6in]{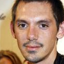} &
\includegraphics[height=0.6in]{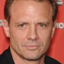} \\
recon &
\includegraphics[height=0.6in]{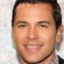} &
\includegraphics[height=0.6in]{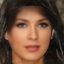} &
\includegraphics[height=0.6in]{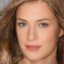} &
\includegraphics[height=0.6in]{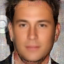} &
\includegraphics[height=0.6in]{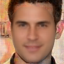} &
\includegraphics[height=0.6in]{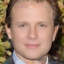} &
\includegraphics[height=0.6in]{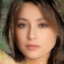} &
\includegraphics[height=0.6in]{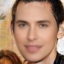} &
\includegraphics[height=0.6in]{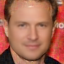} \\
result &
\includegraphics[height=0.6in]{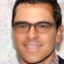} &
\includegraphics[height=0.6in]{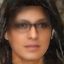} &
\includegraphics[height=0.6in]{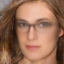} &
\includegraphics[height=0.6in]{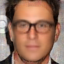} &
\includegraphics[height=0.6in]{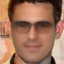} &
\includegraphics[height=0.6in]{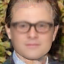} &
\includegraphics[height=0.6in]{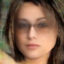} &
\includegraphics[height=0.6in]{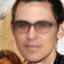} &
\includegraphics[height=0.6in]{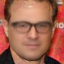} \\~\\

input &
\includegraphics[height=0.6in]{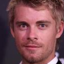} &
\includegraphics[height=0.6in]{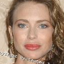} &
\includegraphics[height=0.6in]{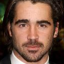} &
\includegraphics[height=0.6in]{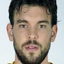} &
\includegraphics[height=0.6in]{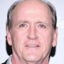} &
\includegraphics[height=0.6in]{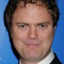} &
\includegraphics[height=0.6in]{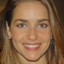} &
\includegraphics[height=0.6in]{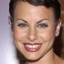} &
\includegraphics[height=0.6in]{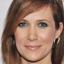} \\
recon &
\includegraphics[height=0.6in]{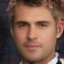} &
\includegraphics[height=0.6in]{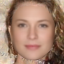} &
\includegraphics[height=0.6in]{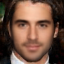} &
\includegraphics[height=0.6in]{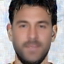} &
\includegraphics[height=0.6in]{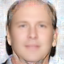} &
\includegraphics[height=0.6in]{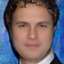} &
\includegraphics[height=0.6in]{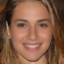} &
\includegraphics[height=0.6in]{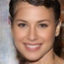} &
\includegraphics[height=0.6in]{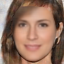} \\
result &
\includegraphics[height=0.6in]{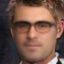} &
\includegraphics[height=0.6in]{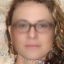} &
\includegraphics[height=0.6in]{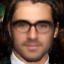} &
\includegraphics[height=0.6in]{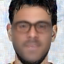} &
\includegraphics[height=0.6in]{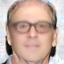} &
\includegraphics[height=0.6in]{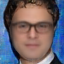} &
\includegraphics[height=0.6in]{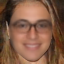} &
\includegraphics[height=0.6in]{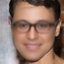} &
\includegraphics[height=0.6in]{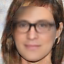} \\
		& (1) & (2) & (3) & (4) & (5) & (6) & (7) & (8) & (9)
	\end{tabular}
	\caption{Adding eye-glasses to the input faces using our proposed method. Edits are computed on the  $Z_{A_i}$ manifold only and $\lambda = 0.02$. }
	\label{fig:supp_eyeglasses}
\end{figure*}

\textbf{Eye-glasses.} Figure~\ref{fig:supp_eyeglasses} shows results of \textit{wear eye-glasses}. For this experiment,  we sample 2000 images from the CelebA~\cite{liu2015deep} of faces \textit{wearing eye-glasses} as $\{\mathbf{x}_p\}$ and 2000 images of faces \textit{not wearing eye-glasses} as $\{\mathbf{x}_n\}$. We compute the edits on the $A_i$ manifold only, with $\lambda=0.02$.

\begin{figure*}
	\centering
	\begin{tabular}{c@{\hspace{0.01in}}c@{\hspace{0.02in}}c@{\hspace{0.02in}}c@{\hspace{0.02in}}c}
		\includegraphics[height=0.6in]{suppimages/glasses/b1/recon/1} &
		\includegraphics[height=0.6in]{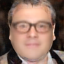} &
		\includegraphics[height=0.6in]{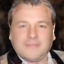} &
		\includegraphics[height=0.6in]{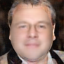} &
		\includegraphics[height=0.6in]{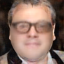} \\
		
		\includegraphics[height=0.6in]{suppimages/glasses/b1/recon/6} &
		\includegraphics[height=0.6in]{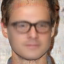} &
		\includegraphics[height=0.6in]{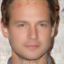} &
		\includegraphics[height=0.6in]{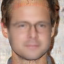} &
		\includegraphics[height=0.6in]{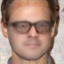} \\
		
		\includegraphics[height=0.6in]{suppimages/glasses/b1/recon/7} &
		\includegraphics[height=0.6in]{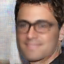} &
		\includegraphics[height=0.6in]{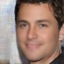} &
		\includegraphics[height=0.6in]{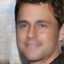} &
		\includegraphics[height=0.6in]{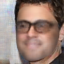} \\
		
		\includegraphics[height=0.6in]{suppimages/glasses/b1/recon/24} &
		\includegraphics[height=0.6in]{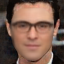} &
		\includegraphics[height=0.6in]{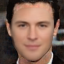} &
		\includegraphics[height=0.6in]{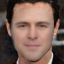} &
		\includegraphics[height=0.6in]{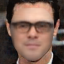} \\
		
		\includegraphics[height=0.6in]{suppimages/glasses/b1/recon/62} &
		\includegraphics[height=0.6in]{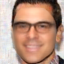} &
		\includegraphics[height=0.6in]{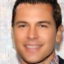} &
		\includegraphics[height=0.6in]{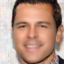} &
		\includegraphics[height=0.6in]{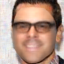} \\

		(a) recon & (b) $Z_{A_i}$ & (c) $Z_{N_i}$ & (d) $Z_{UV}$& (e) all $Z$s
	\end{tabular}
	\caption{Comparison of adding eye-glasses via editing different manifold(s). Editing through $Z_{A_i}$ generates faithful results of wearing eye-glasses (b). Almost no effects show up when manipulating $Z_{N_i}$ (c). Editing through $Z_{UV}$ slightly \textit{aged} the face (d). Editing all three manifolds (e) not only adds eye-glasses, but also changes shape and appearance of the face. $\lambda = 0.02$ for all results in this figure. }
	\label{fig:supp_eyeglasses_manifolds}
\end{figure*}

Since eye-glasses only affect the reflectance in the image, no geometry or warping is associate with this editing, therefore the natural choice of manifold-to-be-edited is $Z_{A_i}$. We show in figure~\ref{fig:supp_eyeglasses_manifolds} a comparison of \textit{adding eye-glasses} via different manifolds (using same the $\lambda = 0.02$). We notice that (1) editing $Z_{N_i}$ (Fig.~\ref{fig:supp_eyeglasses_manifolds}-(c)) almost has no effect on the image; (2) editing $Z_{UV}$ slightly \textit{aged} the face (Fig.~\ref{fig:supp_eyeglasses_manifolds}-(d)), mainly since senior people are more likely to wear eye-glasses; (3) manipulating all the manifolds leads to changes in geometry and appearance (beards, nostrils and shape of noses in Fig.~\ref{fig:supp_eyeglasses_manifolds}-(e)); (4) editing through $Z_{A_i}$ generates faithful results of \textit{wearing eye-glasses} and has little effect on other attributes of the face.

\begin{figure*}
	\centering
	\begin{tabular}{c@{\hspace{0.01in}}c@{\hspace{0.02in}}c@{\hspace{0.02in}}c@{\hspace{0.02in}}c@{\hspace{0.02in}}c@{\hspace{0.02in}}c@{\hspace{0.02in}}c@{\hspace{0.02in}}c@{\hspace{0.02in}}c}
		input &
		\includegraphics[height=0.6in]{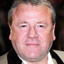} &
		\includegraphics[height=0.6in]{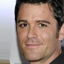} &
		\includegraphics[height=0.6in]{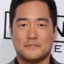} &
		\includegraphics[height=0.6in]{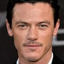} &
		\includegraphics[height=0.6in]{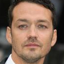} &
		\includegraphics[height=0.6in]{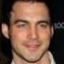} &
		\includegraphics[height=0.6in]{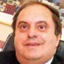} &
		\includegraphics[height=0.6in]{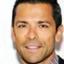} &
		\includegraphics[height=0.6in]{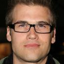} \\
		recon &
		\includegraphics[height=0.6in]{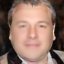} &
		\includegraphics[height=0.6in]{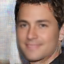} &
		\includegraphics[height=0.6in]{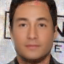} &
		\includegraphics[height=0.6in]{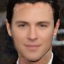} &
		\includegraphics[height=0.6in]{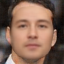} &
		\includegraphics[height=0.6in]{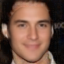} &
		\includegraphics[height=0.6in]{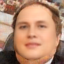} &
		\includegraphics[height=0.6in]{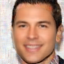} &
		\includegraphics[height=0.6in]{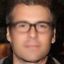} \\
$\lambda = 0.03$ &
\includegraphics[height=0.6in]{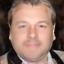} &
\includegraphics[height=0.6in]{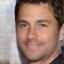} &
\includegraphics[height=0.6in]{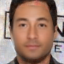} &
\includegraphics[height=0.6in]{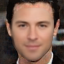} &
\includegraphics[height=0.6in]{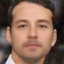} &
\includegraphics[height=0.6in]{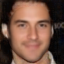} &
\includegraphics[height=0.6in]{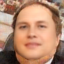} &
\includegraphics[height=0.6in]{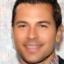} &
\includegraphics[height=0.6in]{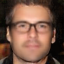} \\
$\lambda = 0.02$ &
\includegraphics[height=0.6in]{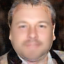} &
\includegraphics[height=0.6in]{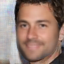} &
\includegraphics[height=0.6in]{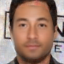} &
\includegraphics[height=0.6in]{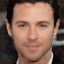} &
\includegraphics[height=0.6in]{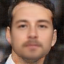} &
\includegraphics[height=0.6in]{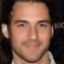} &
\includegraphics[height=0.6in]{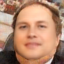} &
\includegraphics[height=0.6in]{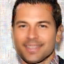} &
\includegraphics[height=0.6in]{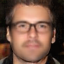} \\
$\lambda = 0.01$ &
\includegraphics[height=0.6in]{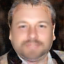} &
\includegraphics[height=0.6in]{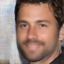} &
\includegraphics[height=0.6in]{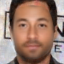} &
\includegraphics[height=0.6in]{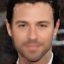} &
\includegraphics[height=0.6in]{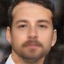} &
\includegraphics[height=0.6in]{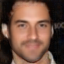} &
\includegraphics[height=0.6in]{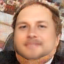} &
\includegraphics[height=0.6in]{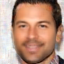} &
\includegraphics[height=0.6in]{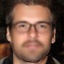} \\~\\
		& (1) & (2) & (3) & (4) & (5) & (6) & (7) & (8) & (9)
\end{tabular}
\caption{\textit{Growing beard} onto the input faces. Edits/traversals are computed on the  manifold $Z_{A_i}$ only with $\lambda = 0.03$, $0.02$, and $0.01$ respectively. }
\label{fig:supp_beard_1}
\end{figure*}

\begin{figure*}
	\centering
	\begin{tabular}{c@{\hspace{0.01in}}c@{\hspace{0.02in}}c@{\hspace{0.02in}}c@{\hspace{0.02in}}c@{\hspace{0.02in}}c@{\hspace{0.02in}}c@{\hspace{0.02in}}c@{\hspace{0.02in}}c@{\hspace{0.02in}}c}
		        input &
		\includegraphics[height=0.6in]{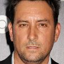} &
		\includegraphics[height=0.6in]{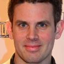} &
		\includegraphics[height=0.6in]{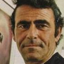} &
		\includegraphics[height=0.6in]{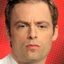} &
		\includegraphics[height=0.6in]{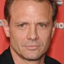} &
		\includegraphics[height=0.6in]{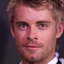} &
		\includegraphics[height=0.6in]{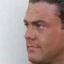} &
		\includegraphics[height=0.6in]{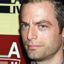} &
		\includegraphics[height=0.6in]{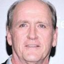} \\
		recon &
		\includegraphics[height=0.6in]{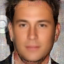} &
		\includegraphics[height=0.6in]{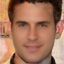} &
		\includegraphics[height=0.6in]{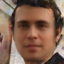} &
		\includegraphics[height=0.6in]{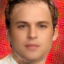} &
		\includegraphics[height=0.6in]{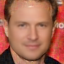} &
		\includegraphics[height=0.6in]{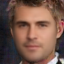} &
		\includegraphics[height=0.6in]{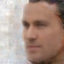} &
		\includegraphics[height=0.6in]{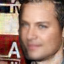} &
		\includegraphics[height=0.6in]{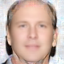} \\
		$\lambda = 0.03$ &
		\includegraphics[height=0.6in]{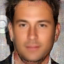} &
		\includegraphics[height=0.6in]{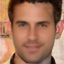} &
		\includegraphics[height=0.6in]{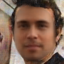} &
		\includegraphics[height=0.6in]{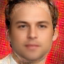} &
		\includegraphics[height=0.6in]{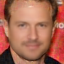} &
		\includegraphics[height=0.6in]{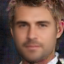} &
		\includegraphics[height=0.6in]{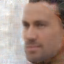} &
		\includegraphics[height=0.6in]{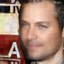} &
		\includegraphics[height=0.6in]{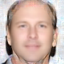} \\
		$\lambda = 0.02$ &
		\includegraphics[height=0.6in]{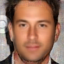} &
		\includegraphics[height=0.6in]{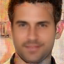} &
		\includegraphics[height=0.6in]{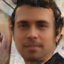} &
		\includegraphics[height=0.6in]{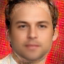} &
		\includegraphics[height=0.6in]{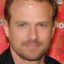} &
		\includegraphics[height=0.6in]{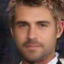} &
		\includegraphics[height=0.6in]{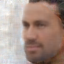} &
		\includegraphics[height=0.6in]{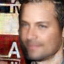} &
		\includegraphics[height=0.6in]{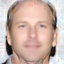} \\
		$\lambda = 0.01$ &
		\includegraphics[height=0.6in]{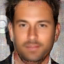} &
		\includegraphics[height=0.6in]{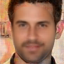} &
		\includegraphics[height=0.6in]{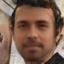} &
		\includegraphics[height=0.6in]{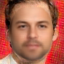} &
		\includegraphics[height=0.6in]{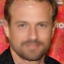} &
		\includegraphics[height=0.6in]{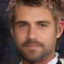} &
		\includegraphics[height=0.6in]{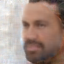} &
		\includegraphics[height=0.6in]{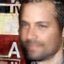} &
		\includegraphics[height=0.6in]{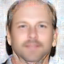} \\
		& (1) & (2) & (3) & (4) & (5) & (6) & (7) & (8) & (9)
	\end{tabular}
	\caption{\textit{Growing beard} onto the input faces. Edits/traversals are computed on the manifold $Z_{A_i}$ only with $\lambda = 0.03$, $0.02$, and $0.01$ respectively. }
	\label{fig:supp_beard_2}
\end{figure*}

\begin{figure*}
	\centering
	\begin{tabular}{c@{\hspace{0.01in}}c@{\hspace{0.02in}}c@{\hspace{0.02in}}c@{\hspace{0.02in}}c@{\hspace{0.02in}}c@{\hspace{0.02in}}c@{\hspace{0.02in}}c@{\hspace{0.02in}}c@{\hspace{0.02in}}c}
		input &
		\includegraphics[height=0.6in]{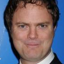} &
		\includegraphics[height=0.6in]{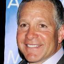} &
		\includegraphics[height=0.6in]{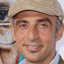} &
		\includegraphics[height=0.6in]{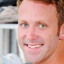} &
		\includegraphics[height=0.6in]{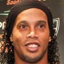} &
		\includegraphics[height=0.6in]{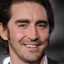} &
		\includegraphics[height=0.6in]{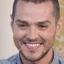} &
		\includegraphics[height=0.6in]{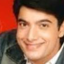} &
		\includegraphics[height=0.6in]{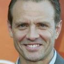} \\
		recon &
		\includegraphics[height=0.6in]{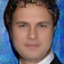} &
		\includegraphics[height=0.6in]{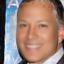} &
		\includegraphics[height=0.6in]{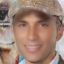} &
		\includegraphics[height=0.6in]{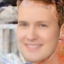} &
		\includegraphics[height=0.6in]{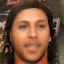} &
		\includegraphics[height=0.6in]{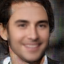} &
		\includegraphics[height=0.6in]{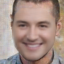} &
		\includegraphics[height=0.6in]{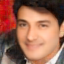} &
		\includegraphics[height=0.6in]{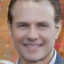} \\
		$\lambda = 0.03$ &
		\includegraphics[height=0.6in]{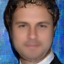} &
		\includegraphics[height=0.6in]{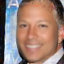} &
		\includegraphics[height=0.6in]{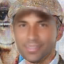} &
		\includegraphics[height=0.6in]{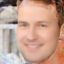} &
		\includegraphics[height=0.6in]{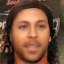} &
		\includegraphics[height=0.6in]{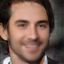} &
		\includegraphics[height=0.6in]{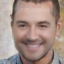} &
		\includegraphics[height=0.6in]{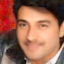} &
		\includegraphics[height=0.6in]{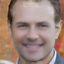} \\
		$\lambda = 0.02$ &
		\includegraphics[height=0.6in]{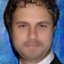} &
		\includegraphics[height=0.6in]{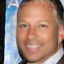} &
		\includegraphics[height=0.6in]{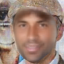} &
		\includegraphics[height=0.6in]{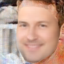} &
		\includegraphics[height=0.6in]{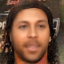} &
		\includegraphics[height=0.6in]{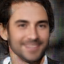} &
		\includegraphics[height=0.6in]{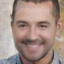} &
		\includegraphics[height=0.6in]{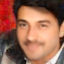} &
		\includegraphics[height=0.6in]{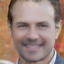} \\
		$\lambda = 0.01$ &
		\includegraphics[height=0.6in]{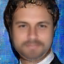} &
		\includegraphics[height=0.6in]{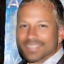} &
		\includegraphics[height=0.6in]{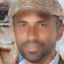} &
		\includegraphics[height=0.6in]{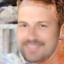} &
		\includegraphics[height=0.6in]{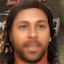} &
		\includegraphics[height=0.6in]{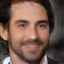} &
		\includegraphics[height=0.6in]{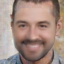} &
		\includegraphics[height=0.6in]{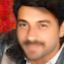} &
		\includegraphics[height=0.6in]{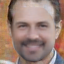} \\
		& (1) & (2) & (3) & (4) & (5) & (6) & (7) & (8) & (9)
	\end{tabular}
	\caption{\textit{Growing beard} onto the input faces. Edits/traversals are computed on the manifold $Z_{A_i}$ only with $\lambda = 0.03$, $0.02$, and $0.01$ respectively. }
	\label{fig:supp_beard_3}
\end{figure*}

\textbf{Beards.} In figures~\ref{fig:supp_beard_1},~\ref{fig:supp_beard_2},~\ref{fig:supp_beard_3}, we present additional results of \textit{grow beard}. For this experiment,  we sample 2000 images from the CelebA~\cite{liu2015deep} of \textit{male} faces \textit{with beard} as $\{\mathbf{x}_p\}$ and 2000 images of \textit{male} faces \textit{without beard} as $\{\mathbf{x}_n\}$. We compute the traversal on the $A_i$ manifold only, with $\lambda = 0.03$, $0.02$, and $0.01$ respectively.

\begin{figure*}
	\centering
	\begin{tabular}{c@{\hspace{0.01in}}c@{\hspace{0.02in}}c@{\hspace{0.02in}}c@{\hspace{0.02in}}c@{\hspace{0.02in}}c@{\hspace{0.02in}}c@{\hspace{0.02in}}c@{\hspace{0.02in}}c@{\hspace{0.02in}}c}
		input &
		\includegraphics[height=0.6in]{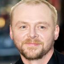} &
		\includegraphics[height=0.6in]{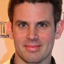} &
		\includegraphics[height=0.6in]{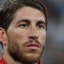} &
		\includegraphics[height=0.6in]{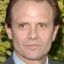} &
		\includegraphics[height=0.6in]{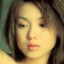} &
		\includegraphics[height=0.6in]{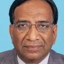} &
		\includegraphics[height=0.6in]{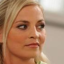} &
		\includegraphics[height=0.6in]{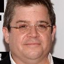} &
		\includegraphics[height=0.6in]{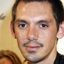} \\
		recon &
		\includegraphics[height=0.6in]{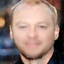} &
		\includegraphics[height=0.6in]{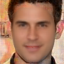} &
		\includegraphics[height=0.6in]{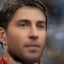} &
		\includegraphics[height=0.6in]{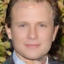} &
		\includegraphics[height=0.6in]{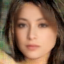} &
		\includegraphics[height=0.6in]{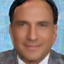} &
		\includegraphics[height=0.6in]{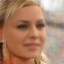} &
		\includegraphics[height=0.6in]{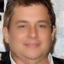} &
		\includegraphics[height=0.6in]{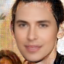} \\
		$\lambda = 0.05$ &
		\includegraphics[height=0.6in]{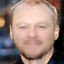} &
		\includegraphics[height=0.6in]{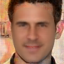} &
		\includegraphics[height=0.6in]{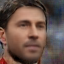} &
		\includegraphics[height=0.6in]{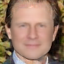} &
		\includegraphics[height=0.6in]{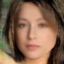} &
		\includegraphics[height=0.6in]{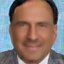} &
		\includegraphics[height=0.6in]{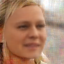} &
		\includegraphics[height=0.6in]{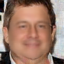} &
		\includegraphics[height=0.6in]{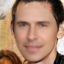} \\
		$\lambda = 0.03$ &
		\includegraphics[height=0.6in]{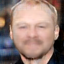} &
		\includegraphics[height=0.6in]{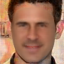} &
		\includegraphics[height=0.6in]{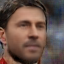} &
		\includegraphics[height=0.6in]{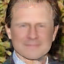} &
		\includegraphics[height=0.6in]{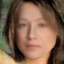} &
		\includegraphics[height=0.6in]{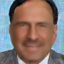} &
		\includegraphics[height=0.6in]{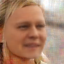} &
		\includegraphics[height=0.6in]{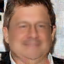} &
		\includegraphics[height=0.6in]{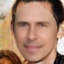} \\
		$\lambda = 0.02$ &
		\includegraphics[height=0.6in]{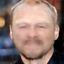} &
		\includegraphics[height=0.6in]{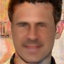} &
		\includegraphics[height=0.6in]{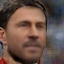} &
		\includegraphics[height=0.6in]{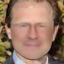} &
		\includegraphics[height=0.6in]{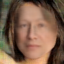} &
		\includegraphics[height=0.6in]{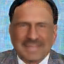} &
		\includegraphics[height=0.6in]{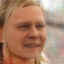} &
		\includegraphics[height=0.6in]{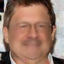} &
		\includegraphics[height=0.6in]{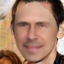} \\
		& (1) & (2) & (3) & (4) & (5) & (6) & (7) & (8) & (9)
	\end{tabular}
	\caption{\textit{Aging}. Edits/traversals are computeg on manifolds $Z_{A_i}$,  $Z_{N_i}$, and  $Z_{UV}$, with $\lambda = 0.05$, $0.03$, and $0.02$ respectively. }
	\label{fig:supp_young_1}
\end{figure*}

\begin{figure*}
	\centering
	\begin{tabular}{c@{\hspace{0.01in}}c@{\hspace{0.02in}}c@{\hspace{0.02in}}c@{\hspace{0.02in}}c@{\hspace{0.02in}}c@{\hspace{0.02in}}c@{\hspace{0.02in}}c@{\hspace{0.02in}}c@{\hspace{0.02in}}c}
		input &
		\includegraphics[height=0.6in]{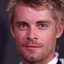} &
		\includegraphics[height=0.6in]{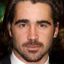} &
		\includegraphics[height=0.6in]{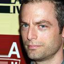} &
		\includegraphics[height=0.6in]{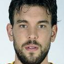} &
		\includegraphics[height=0.6in]{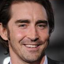} &
		\includegraphics[height=0.6in]{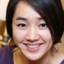} &
		\includegraphics[height=0.6in]{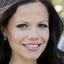} &
		\includegraphics[height=0.6in]{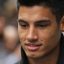} &
		\includegraphics[height=0.6in]{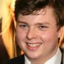} \\
		recon &
		\includegraphics[height=0.6in]{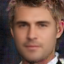} &
		\includegraphics[height=0.6in]{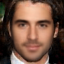} &
		\includegraphics[height=0.6in]{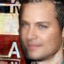} &
		\includegraphics[height=0.6in]{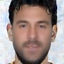} &
		\includegraphics[height=0.6in]{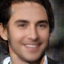} &
		\includegraphics[height=0.6in]{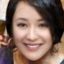} &
		\includegraphics[height=0.6in]{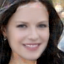} &
		\includegraphics[height=0.6in]{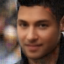} &
		\includegraphics[height=0.6in]{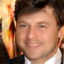} \\
		$\lambda = 0.05$ &
		\includegraphics[height=0.6in]{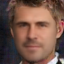} &
		\includegraphics[height=0.6in]{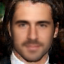} &
		\includegraphics[height=0.6in]{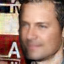} &
		\includegraphics[height=0.6in]{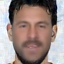} &
		\includegraphics[height=0.6in]{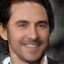} &
		\includegraphics[height=0.6in]{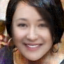} &
		\includegraphics[height=0.6in]{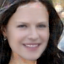} &
		\includegraphics[height=0.6in]{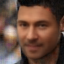} &
		\includegraphics[height=0.6in]{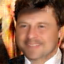} \\
		$\lambda = 0.03$ &
		\includegraphics[height=0.6in]{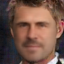} &
		\includegraphics[height=0.6in]{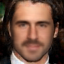} &
		\includegraphics[height=0.6in]{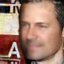} &
		\includegraphics[height=0.6in]{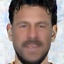} &
		\includegraphics[height=0.6in]{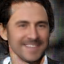} &
		\includegraphics[height=0.6in]{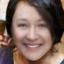} &
		\includegraphics[height=0.6in]{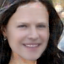} &
		\includegraphics[height=0.6in]{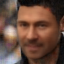} &
		\includegraphics[height=0.6in]{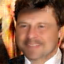} \\
		$\lambda = 0.02$ &
		\includegraphics[height=0.6in]{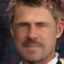} &
		\includegraphics[height=0.6in]{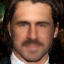} &
		\includegraphics[height=0.6in]{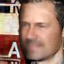} &
		\includegraphics[height=0.6in]{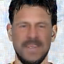} &
		\includegraphics[height=0.6in]{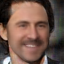} &
		\includegraphics[height=0.6in]{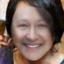} &
		\includegraphics[height=0.6in]{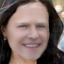} &
		\includegraphics[height=0.6in]{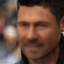} &
		\includegraphics[height=0.6in]{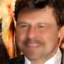} \\
		& (1) & (2) & (3) & (4) & (5) & (6) & (7) & (8) & (9)
	\end{tabular}
	\caption{\textit{Aging}. Edits/traversals are computed on manifolds $Z_{A_i}$,  $Z_{N_i}$, and  $Z_{UV}$, with $\lambda = 0.05$, $0.03$, and $0.02$ respectively. }
	\label{fig:supp_young_2}
\end{figure*}

\textbf{Aging.} Figures~\ref{fig:supp_young_1},~\ref{fig:supp_young_2}, show additional results of \textit{aging}. We compute the traversal on manifolds $Z_{A_i}$,  $Z_{N_i}$, and  $Z_{UV}$, with $\lambda = 0.05$, $0.03$, and $0.02$ respectively.

\begin{figure*}
	\centering
	\begin{tabular}{c@{\hspace{0.01in}}c@{\hspace{0.02in}}c@{\hspace{0.02in}}c@{\hspace{0.02in}}c@{\hspace{0.02in}}c@{\hspace{0.02in}}c@{\hspace{0.02in}}c@{\hspace{0.02in}}c@{\hspace{0.02in}}c}
		input &
		\includegraphics[height=0.6in]{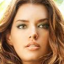} &
		\includegraphics[height=0.6in]{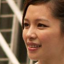} &
		\includegraphics[height=0.6in]{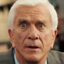} &
		\includegraphics[height=0.6in]{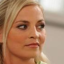} &
		\includegraphics[height=0.6in]{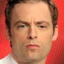} &
		\includegraphics[height=0.6in]{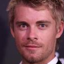} &
		\includegraphics[height=0.6in]{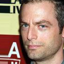} &
		\includegraphics[height=0.6in]{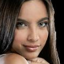} &
		\includegraphics[height=0.6in]{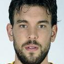} \\
		recon &
		\includegraphics[height=0.6in]{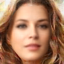} &
		\includegraphics[height=0.6in]{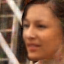} &
		\includegraphics[height=0.6in]{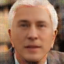} &
		\includegraphics[height=0.6in]{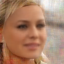} &
		\includegraphics[height=0.6in]{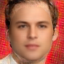} &
		\includegraphics[height=0.6in]{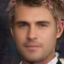} &
		\includegraphics[height=0.6in]{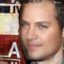} &
		\includegraphics[height=0.6in]{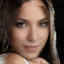} &
		\includegraphics[height=0.6in]{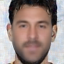} \\
		$\lambda = 0.05$ &
		\includegraphics[height=0.6in]{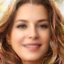} &
		\includegraphics[height=0.6in]{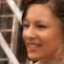} &
		\includegraphics[height=0.6in]{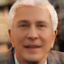} &
		\includegraphics[height=0.6in]{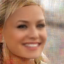} &
		\includegraphics[height=0.6in]{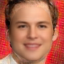} &
		\includegraphics[height=0.6in]{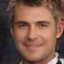} &
		\includegraphics[height=0.6in]{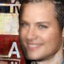} &
		\includegraphics[height=0.6in]{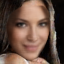} &
		\includegraphics[height=0.6in]{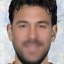} \\
		$\lambda = 0.03$ &
		\includegraphics[height=0.6in]{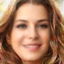} &
		\includegraphics[height=0.6in]{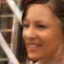} &
		\includegraphics[height=0.6in]{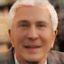} &
		\includegraphics[height=0.6in]{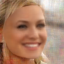} &
		\includegraphics[height=0.6in]{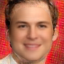} &
		\includegraphics[height=0.6in]{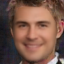} &
		\includegraphics[height=0.6in]{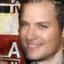} &
		\includegraphics[height=0.6in]{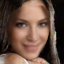} &
		\includegraphics[height=0.6in]{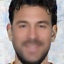} \\
		$\lambda = 0.02$ &
		\includegraphics[height=0.6in]{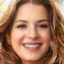} &
		\includegraphics[height=0.6in]{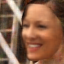} &
		\includegraphics[height=0.6in]{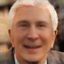} &
		\includegraphics[height=0.6in]{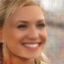} &
		\includegraphics[height=0.6in]{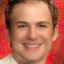} &
		\includegraphics[height=0.6in]{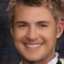} &
		\includegraphics[height=0.6in]{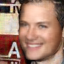} &
		\includegraphics[height=0.6in]{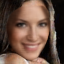} &
		\includegraphics[height=0.6in]{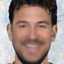} \\
		& (1) & (2) & (3) & (4) & (5) & (6) & (7) & (8) & (9)
	\end{tabular}
	\caption{\textit{Smiling}. Edits/traversals are computed on the manifolds $Z_{N_i}$ and  $Z_{UV}$, with $\lambda = 0.07$, $0.05$, and $0.03$ respectively. }
	\label{fig:supp_smile_1}
\end{figure*}

\begin{figure*}
	\centering
	\begin{tabular}{c@{\hspace{0.01in}}c@{\hspace{0.02in}}c@{\hspace{0.02in}}c@{\hspace{0.02in}}c@{\hspace{0.02in}}c@{\hspace{0.02in}}c@{\hspace{0.02in}}c@{\hspace{0.02in}}c@{\hspace{0.02in}}c}
		input &
		\includegraphics[height=0.6in]{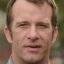} &
		\includegraphics[height=0.6in]{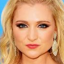} &
		\includegraphics[height=0.6in]{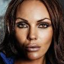} &
		\includegraphics[height=0.6in]{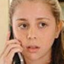} &
		\includegraphics[height=0.6in]{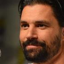} &
		\includegraphics[height=0.6in]{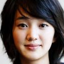} &
		\includegraphics[height=0.6in]{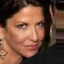} &
		\includegraphics[height=0.6in]{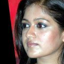} &
		\includegraphics[height=0.6in]{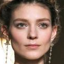} \\
		recon &
		\includegraphics[height=0.6in]{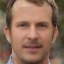} &
		\includegraphics[height=0.6in]{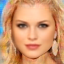} &
		\includegraphics[height=0.6in]{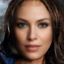} &
		\includegraphics[height=0.6in]{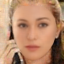} &
		\includegraphics[height=0.6in]{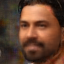} &
		\includegraphics[height=0.6in]{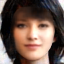} &
		\includegraphics[height=0.6in]{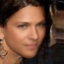} &
		\includegraphics[height=0.6in]{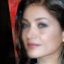} &
		\includegraphics[height=0.6in]{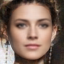} \\
		$\lambda = 0.05$ &
		\includegraphics[height=0.6in]{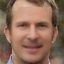} &
		\includegraphics[height=0.6in]{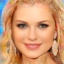} &
		\includegraphics[height=0.6in]{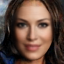} &
		\includegraphics[height=0.6in]{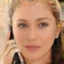} &
		\includegraphics[height=0.6in]{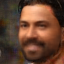} &
		\includegraphics[height=0.6in]{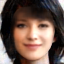} &
		\includegraphics[height=0.6in]{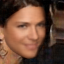} &
		\includegraphics[height=0.6in]{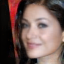} &
		\includegraphics[height=0.6in]{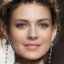} \\
		$\lambda = 0.03$ &
		\includegraphics[height=0.6in]{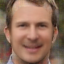} &
		\includegraphics[height=0.6in]{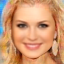} &
		\includegraphics[height=0.6in]{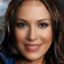} &
		\includegraphics[height=0.6in]{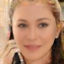} &
		\includegraphics[height=0.6in]{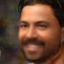} &
		\includegraphics[height=0.6in]{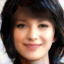} &
		\includegraphics[height=0.6in]{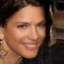} &
		\includegraphics[height=0.6in]{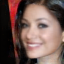} &
		\includegraphics[height=0.6in]{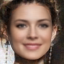} \\
		$\lambda = 0.02$ &
		\includegraphics[height=0.6in]{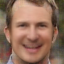} &
		\includegraphics[height=0.6in]{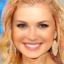} &
		\includegraphics[height=0.6in]{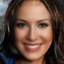} &
		\includegraphics[height=0.6in]{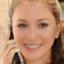} &
		\includegraphics[height=0.6in]{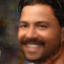} &
		\includegraphics[height=0.6in]{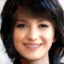} &
		\includegraphics[height=0.6in]{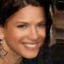} &
		\includegraphics[height=0.6in]{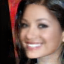} &
		\includegraphics[height=0.6in]{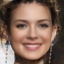} \\
		& (1) & (2) & (3) & (4) & (5) & (6) & (7) & (8) & (9)
	\end{tabular}
	\caption{\textit{Smiling}. Edits/traversals are computed on manifolds $Z_{N_i}$ and  $Z_{UV}$, with $\lambda = 0.07$, $0.05$, and $0.03$ respectively. }
	\label{fig:supp_smile_2}
\end{figure*}

\textbf{Smiling.}  Figures~\ref{fig:supp_smile_1},~\ref{fig:supp_smile_2} show additional results of \textit{smiling} as described in Section 5.2 of the main paper. The data is as described in the main paper. We compute the traversal on manifolds $Z_{N_i}$, and  $Z_{UV}$, with $\lambda = 0.07$, $0.05$, and $0.03$ respectively.

\begin{figure}
	\centering
	\begin{tabular}{c@{\hspace{0.05in}}c@{\hspace{0.05in}}c@{\hspace{0.05in}}c@{\hspace{0.05in}}c}
		
		\includegraphics[height=0.6in]{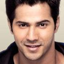} &
		\includegraphics[height=0.6in]{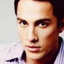} &
		\includegraphics[height=0.6in]{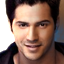} &
		\includegraphics[height=0.6in]{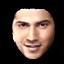} &
		\includegraphics[height=0.6in]{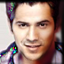} \\
		&
		\includegraphics[height=0.6in]{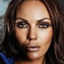} &
		\includegraphics[height=0.6in]{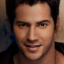} &
		\includegraphics[height=0.6in]{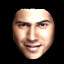} &
		\includegraphics[height=0.6in]{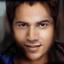} \\
		(a) target & (b) source & (c) ours & (d) SH & (e) EPF 
	\end{tabular}
	\caption{Comparison of re-lighting with spherical harmonics-based radiance maps (SH)~\protect\cite{wen2003face} and edge-preserving filters (EPF)~\protect\cite{chen2011face}.}
	\label{fig:lighting_transfer_compare}
\end{figure}

\begin{figure*}
	\centering
	\begin{tabular}{c@{\hspace{0.01in}}c@{\hspace{0.02in}}c@{\hspace{0.02in}}c@{\hspace{0.02in}}c@{\hspace{0.02in}}c@{\hspace{0.02in}}c@{\hspace{0.02in}}c@{\hspace{0.02in}}c@{\hspace{0.02in}}c}
		
		source & 
		\includegraphics[height=0.6in]{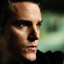}& & & & & & & &\\
		$S^{\text{source}}$ &
		\includegraphics[height=0.6in]{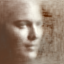} & & & & & & & &\\~\\
		target &
		\includegraphics[height=0.6in]{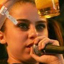} &
		\includegraphics[height=0.6in]{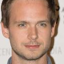} &
		\includegraphics[height=0.6in]{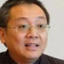} &
		\includegraphics[height=0.6in]{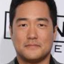} &
		\includegraphics[height=0.6in]{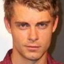} &
		\includegraphics[height=0.6in]{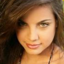} &
		\includegraphics[height=0.6in]{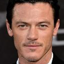} &
		\includegraphics[height=0.6in]{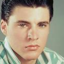} &
		\includegraphics[height=0.6in]{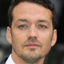} \\
		$S^{\text{target}}$ &
		\includegraphics[height=0.6in]{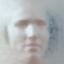} &
		\includegraphics[height=0.6in]{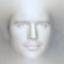} &
		\includegraphics[height=0.6in]{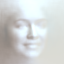} &
		\includegraphics[height=0.6in]{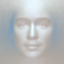} &
		\includegraphics[height=0.6in]{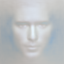} &
		\includegraphics[height=0.6in]{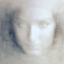} &
		\includegraphics[height=0.6in]{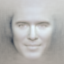} &
		\includegraphics[height=0.6in]{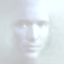} &
		\includegraphics[height=0.6in]{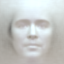} \\
		$S^{\text{transfer}}$ &
		\includegraphics[height=0.6in]{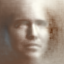} &
		\includegraphics[height=0.6in]{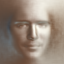} &
		\includegraphics[height=0.6in]{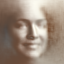} &
		\includegraphics[height=0.6in]{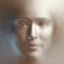} &
		\includegraphics[height=0.6in]{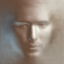} &
		\includegraphics[height=0.6in]{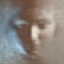} &
		\includegraphics[height=0.6in]{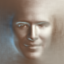} &
		\includegraphics[height=0.6in]{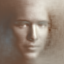} &
		\includegraphics[height=0.6in]{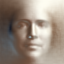} \\
		transfer &
		\includegraphics[height=0.6in]{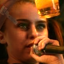} &
		\includegraphics[height=0.6in]{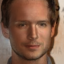} &
		\includegraphics[height=0.6in]{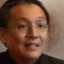} &
		\includegraphics[height=0.6in]{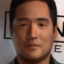} &
		\includegraphics[height=0.6in]{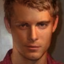} &
		\includegraphics[height=0.6in]{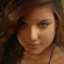} &
		\includegraphics[height=0.6in]{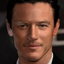} &
		\includegraphics[height=0.6in]{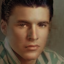} &
		\includegraphics[height=0.6in]{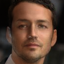} \\~\\
				target &
		\includegraphics[height=0.6in]{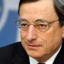} &
		\includegraphics[height=0.6in]{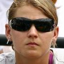} &
		\includegraphics[height=0.6in]{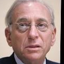} &
		\includegraphics[height=0.6in]{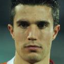} &
		\includegraphics[height=0.6in]{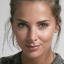} &
		\includegraphics[height=0.6in]{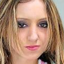} &
		\includegraphics[height=0.6in]{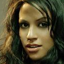} &
		\includegraphics[height=0.6in]{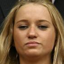} &
		\includegraphics[height=0.6in]{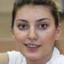} \\
		$S^{\text{target}}$ &
		\includegraphics[height=0.6in]{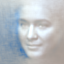} &
		\includegraphics[height=0.6in]{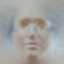} &
		\includegraphics[height=0.6in]{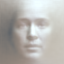} &
		\includegraphics[height=0.6in]{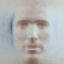} &
		\includegraphics[height=0.6in]{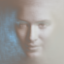} &
		\includegraphics[height=0.6in]{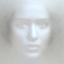} &
		\includegraphics[height=0.6in]{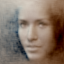} &
		\includegraphics[height=0.6in]{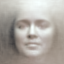} &
		\includegraphics[height=0.6in]{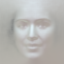} \\
		$S^{\text{transfer}}$ &
		\includegraphics[height=0.6in]{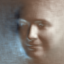} &
		\includegraphics[height=0.6in]{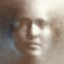} &
		\includegraphics[height=0.6in]{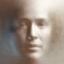} &
		\includegraphics[height=0.6in]{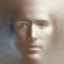} &
		\includegraphics[height=0.6in]{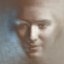} &
		\includegraphics[height=0.6in]{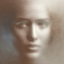} &
		\includegraphics[height=0.6in]{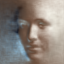} &
		\includegraphics[height=0.6in]{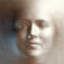} &
		\includegraphics[height=0.6in]{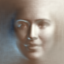} \\
		transfer &
		\includegraphics[height=0.6in]{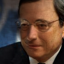} &
		\includegraphics[height=0.6in]{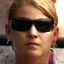} &
		\includegraphics[height=0.6in]{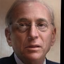} &
		\includegraphics[height=0.6in]{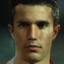} &
		\includegraphics[height=0.6in]{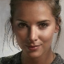} &
		\includegraphics[height=0.6in]{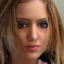} &
		\includegraphics[height=0.6in]{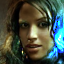} &
		\includegraphics[height=0.6in]{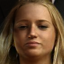} &
		\includegraphics[height=0.6in]{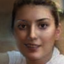} \\
		& (1) & (2) & (3) & (4) & (5) & (6) & (7) & (8) & (9)
	\end{tabular}
	\caption{\textit{Relighting}. Transfer of the lighting condition from the  source image to target images using the method described in Section 5.3 of the main paper.}
	\label{fig:relit_1}
\end{figure*}

\textbf{Relighting.} We present additional relighting results in Fig.~\ref*{fig:relit_1}.  In addition, in Fig.~\ref{fig:lighting_transfer_compare}, we provide comparisons to two previous techniques for re-lighting~\cite{wen2003face,chen2011face}. Our results apply to the full face, capture the target lighting, and have fewer artifacts. Our general face editing technique is able to produce results that are qualitatively similar, or even better than other techniques specifically designed for this particular task.

\clearpage

\clearpage

\clearpage

\clearpage

\clearpage

\clearpage

\clearpage

\clearpage

\clearpage

\clearpage

\end{document}